%% file: iclr2026_conference.tex
\documentclass{article} 
\usepackage{iclr2026_conference,times}

\input{math_commands.tex}

\usepackage[utf8]{inputenc} 
\usepackage[T1]{fontenc}    
\usepackage{hyperref}       
\usepackage{url}            
\usepackage{booktabs}       
\usepackage{amsfonts}       
\usepackage{nicefrac}       
\usepackage{microtype}      
\usepackage{xcolor}         
\usepackage{soul}
\usepackage{wrapfig} 

\usepackage{adjustbox}
\usepackage{colortbl} 
\usepackage{multirow} 
\usepackage{makecell} 
\usepackage{amsmath} 
\usepackage{xspace} 
\usepackage{lipsum} 
\usepackage{enumitem} 
\usepackage{duckuments} 
\usepackage{multirow}
\usepackage[most]{tcolorbox} 
\usepackage{pifont} 
\usepackage{fontawesome5}
\usepackage{subcaption}
\usepackage{longtable}
\usepackage{fancyvrb}
\usepackage{tabularx}
\usepackage{ragged2e}
\tcbuselibrary{skins, breakable} 
\usepackage[scaled=0.88]{zi4}
\usepackage[scaled=0.88]{beramono}
\newcommand{\prettyfont}{\fontfamily{zi4}\selectfont}
\newcommand{\pagefont}{\fontfamily{fvm}\selectfont}
\newcommand{\promptfont}{\fontfamily{bch}\selectfont}
\newcommand{\prettyurl}[1]{%
  {\def\UrlFont{\prettyfont}\url{#1}}%
}

\input{macro}

\title{
\system{}: Learning to Decompile 
\\
Symbolic Music to Programs
}

\author{Yewon Kim{\quad}Apurva Gandhi{\quad}David Chung{\quad}Graham Neubig{\quad}Chris Donahue\\
Carnegie Mellon University 
}

\iclrfinalcopy 

\begin{document}

\maketitle


\begin{abstract}
Musical performance involves 
executing a set of high-level musical instructions, 
yet recovering those instructions from the performance is a challenging inverse problem. 
We present \system{}, 
a post-training framework for 
\emph{symbolic music decompilation}: the task of 
recovering executable, editable music programs 
from symbolic music.
We instantiate the task as 
MIDI-to-\Strudel{} decompilation, 
where the model takes symbolic MIDI as input and produces a program in \Strudel{}, a music programming language, that reconstructs the input when executed. 
The task poses two challenges: 
\Strudel{} is a low-resource language 
with little naturally paired MIDI--code data,
and optimizing faithful reconstruction of MIDI alone 
can collapse to unreadable note-by-note transliteration.
We address these challenges in two stages.
First, we construct \dataset{}, 
a synthetic corpus of paired \Strudel{} programs and rendered MIDI, 
and use it for supervised fine-tuning. 
Second, we refine the model with 
reinforcement learning on unpaired MIDI, 
optimizing rewards for both MIDI reconstruction faithfulness and code readability. 
Our evaluation across synthetic and real-world MIDI benchmarks shows that  
\system{} achieves substantially higher MIDI reconstruction faithfulness than closed-source  LLMs while producing more readable and diverse code than the heuristic converter.

\vspace{-0.08in}
\end{abstract}

\input{sections/1_introduction}
\input{sections/2_preliminaries}
\input{sections/3_method}
\input{sections/4_experiments}
\input{sections/5_related}

\input{sections/6_conclusion}

\section*{Acknowledgment}
We thank 
Michael Freeman for sharing his ModArchive-to-MIDI conversion code and resources, which made our genre-wise evaluation of \system{} possible.
We also thank Sihyun Yu for insightful discussions and feedback throughout the project. 
This work was supported by the Humanities and AI Virtual Institute (HAVI) at Schmidt Sciences. Apurva Gandhi is supported by the Amazon AI PhD Fellowship.

\newpage
\bibliography{iclr2026_conference}
\bibliographystyle{iclr2026_conference}

\newpage
\input{sections/appendix}

\end{document}

%% file: math_commands.tex
\usepackage{amsmath,amsfonts,bm}

\def\eqref#1{equation~\ref{#1}}

\def\1{\bm{1}}

\DeclareMathAlphabet{\mathsfit}{\encodingdefault}{\sfdefault}{m}{sl}
\SetMathAlphabet{\mathsfit}{bold}{\encodingdefault}{\sfdefault}{bx}{n}



%% file: macro.tex
\definecolor{cornellred}{rgb}{0.7, 0.11, 0.11}
\definecolor{yblue}{HTML}{5793AA}
\definecolor{yviolet}{HTML}{564A94}
\definecolor{ypaleblue}{HTML}{CEDDFA}
\definecolor{yred}{HTML}{AF3E63}
\definecolor{ywafer}{HTML}{E5D8D2}
\definecolor{ydarkblue}{HTML}{22254E}
\definecolor{lightgray}{HTML}{ECECEC}

\definecolor{ylightpurple}{HTML}{F6F5F8}
\definecolor{ydarkgrayazure}{HTML}{1A2125}
\definecolor{ydarkblue}{HTML}{1D1F42}
\definecolor{glightpurple}{HTML}{EDE7F7}
\definecolor{glightpurple2}{HTML}{d1d1e7}

\definecolor{aliceblue}{rgb}{0.91, 0.94, 0.97}
\definecolor{darkblue}{rgb}{0.83, 0.89, 0.97}
\definecolor{mutedviolet}{HTML}{EEDAFF} 
\definecolor{mutedpink}{HTML}{FFD9D9} 
\definecolor{mutedblue}{HTML}{CCE5FF} 
\definecolor{mutedgreen}{HTML}{CEF8C2} 
\definecolor{forestgreen}{rgb}{0.13, 0.55, 0.13}
\definecolor{hookergreen}{rgb}{0.0, 0.44, 0.0}
\definecolor{cozygreen}{RGB}{170, 210, 160} 
\definecolor{vviolet}{HTML}{676aa4} 
\definecolor{vpink}{HTML}{bb7699} 
\definecolor{vblue}{HTML}{618eb4} 
\definecolor{vgreen}{HTML}{367660}

\newcommand{\Strudel}{Strudel}
\newcommand{\system}{\textsc{Decomposer}}
\newcommand{\dataset}{\textsc{Strudel-Synth}}
\newcommand{\claude}{Claude-Opus-4.6}
\newcommand{\gemini}{Gemini-3.5-Flash}
\newcommand{\gpt}{GPT-5.5}

\hypersetup{
  linkcolor = ydarkblue,
  citecolor  = yblue,
  colorlinks = true,
  urlcolor = yviolet
}

\newcommand\eg{e.g.,~}
\newcommand\ie{i.e.,~}

\newcommand\blfootnote[1]{%
  \begingroup
  \renewcommand\thefootnote{}\footnote{#1}%
  \addtocounter{footnote}{-1}%
  \endgroup
}

\newcommand{\promptsection}[1]{%
  \vspace{0.6em}
  \noindent\textbf{#1}\par
  \vspace{0.15em}
}

\newcommand{\promptsubsection}[1]{%
  \vspace{0.35em}
  \noindent\textit{#1}\par
  \vspace{0.05em}
}

\newcommand{\promptplaceholder}[1]{%
  \smallskip
  \noindent\hspace*{1.5em}{\texttt{#1}}%
  \smallskip
}

%% file: sections/1_introduction.tex
\section{Introduction}
\vspace{-0.03in}
\label{intro}


\input{figures/fig_music_decompilation}
Decompilation, 
the task of 
generating 
an editable, re-executable program 
from a rendered artifact, 
has emerged as a central problem across diverse modalities:
binaries to source code~\citep{tan2024llm4decompile, zou2025dlift},
raster images to SVG programs~\citep{juan2025rlrfsvg, rodriguez2024starvector},
UI screenshots to frontend code~\citep{vanita2019sketch2code, wu2025ui2code},
and 3D shapes to CAD scripts~\citep{kolodiazhnyi2026cadrille, chen2025cadcrafter}.
The motivation is shared across these domains: 
rendered artifacts are often 
the only representation available in practice,
yet downstream use---analysis, modification, integration with existing tooling---benefits from a programmatic form that
human users can read, modify, and extend~\citep{rodriguez2024starvector, tan2024llm4decompile, zou2025dlift}.
A direct 
consequence is that
functional equivalence alone is insufficient: 
a recovered program that
reproduces the artifact but is unreadable defeats the purpose of 
decompilation~\citep{bnw, scalabrino2018readability}.
Effective decompilation thus 
requires jointly optimizing 
\emph{faithfulness} (the program reproduces the input artifact)
and \emph{readability} (the program is easy to read and edit).
\blfootnote{\hfill \textbf{Project page:} {\pagefont\href{https://yewon-kim.com/decomposer}{yewon-kim.com/decomposer}}}


We extend this paradigm to the music modality and introduce 
\emph{symbolic music decompilation}: 
generating code in a music programming language~\citep{chuck, puredata, strudelpaper, sonicpi, tidalcycle, supercollider} corresponding to a symbolic music input. 
We instantiate the paradigm here through a MIDI-to-\Strudel{}~\citep{strudelpaper} task, 
where \Strudel{} is a domain-specific language (DSL) for algorithmic composition~\citep{dean2018oxford} and live coding~\citep{collins2003live, wang2004fly} with JavaScript syntax. 
Unlike MIDI, which represents music as a low-level sequence of note events,
\Strudel{} represents music through high-level pattern expressions and
programmatic operators that can encode repetition, layering, and temporal
structure~(Figure~\ref{fig:music_decompilation}; see Appendix~\ref{app:strudel:examples} for more examples). 
Recovering \Strudel{} code from MIDI therefore does more than reproduce the
input notes: it exposes latent musical structure that MIDI flattens into individual
events, produces code that is readable and directly editable, and places music
generation and editing in a code modality well suited to modern language model
tooling.

\input{figures/fig_system}

MIDI-to-\Strudel{} decompilation is challenging for both task-level and
data-level reasons. 
At the task level, 
faithfulness alone does not define a
useful decompilation: 
a heuristic converter can achieve
near-perfect rendered-MIDI faithfulness~\citep{midi_to_strudel},
yet it does so by emitting verbose note-level code 
that ignores the abstractions
that make \Strudel{} useful to humans. 
In contrast, 
frontier LLMs~\citep{openai2026gpt55, anthropic2026opus46, google2026gemini35flash} often generate readable \Strudel{} programs, but 
their rendered MIDI is not faithful to the input.
At the data level, this mapping is hard
to learn because naturally paired \mbox{(MIDI, \Strudel{})} 
examples are not available.
Moreover, \Strudel{} itself is low-resource: 
our initial crawl
found only $688$ non-trivial human-authored programs,
and open-weight LLMs rarely produce 
executable \Strudel{} without adaptation, 
making direct execution-based optimization difficult.

\paragraph{Our approach.}
We present \system{}, a framework for post-training LLMs to decompile MIDI 
into \Strudel{} code~(Figure~\ref{fig:main}). 
The framework consists of a two-stage training pipeline. 
First, supervised fine-tuning (SFT) teaches a model to
produce valid and idiomatic \Strudel{} code, 
providing a warm start for on-policy learning. 
Second, reinforcement learning (RL) optimizes the
decompilation objective directly. 
For each MIDI input, 
the model samples candidate programs, 
executes them with the \Strudel{} runtime, 
and receives a composite reward that measures both rendered-MIDI 
faithfulness and code-level readability.
Because this reward depends only on the input MIDI and the generated program,
the RL stage can train on any unpaired MIDI at scale 
without requiring reference \Strudel{} code. 
To supply the paired examples needed for the SFT warm start,
we additionally construct and release \dataset{}, 
a synthetic \mbox{(MIDI, \Strudel{})} corpus 
obtained by generating \Strudel{} programs with a frontier LLM and rendering each program to MIDI.

Our experiments show that \system{} substantially improves MIDI-to-\Strudel{} decompilation, 
producing programs that better reconstruct the input MIDI while preserving readable code structure.
On both \dataset{} and real-world LMD~\citep{lakhmidi}, \system{} produces markedly more faithful renderings than frontier LLMs~\citep{anthropic2026opus46, google2026gemini35flash, openai2026gpt55}, 
which often generate plausible-looking but poorly grounded \Strudel{} code. 
At the same time, 
unlike a heuristic converter~\citep{midi_to_strudel} that
achieves high faithfulness by emitting verbose note-level transcriptions, 
\system{} preserves readable program structure comparable to frontier LLMs.

We highlight the main contributions of this paper below:

\begin{itemize}[leftmargin=*,itemsep=1mm]
\item 
We introduce \emph{symbolic music decompilation}, the task of recovering
executable, editable music-programming-language code from symbolic music input.

\item 
We propose \system{}, a two-stage post-training framework that combines
supervised fine-tuning with execution-based reinforcement learning for
symbolic music decompilation.

\item We release \dataset{}, a synthetic corpus of $21{,}174$ 
(\Strudel{}, MIDI) pairs constructed by frontier LLM generation of \Strudel{} code, followed by \Strudel{}-to-MIDI code execution.

\item We find that an 8B open-weight model trained with our pipeline substantially outperforms closed-weight frontier LLMs on decompilation faithfulness, while retaining comparable readability.


\end{itemize}

%% file: figures/fig_music_decompilation.tex
\begin{wrapfigure}{r}{0.5\textwidth}
\vspace{-0.2in}
    \includegraphics[width=\linewidth]{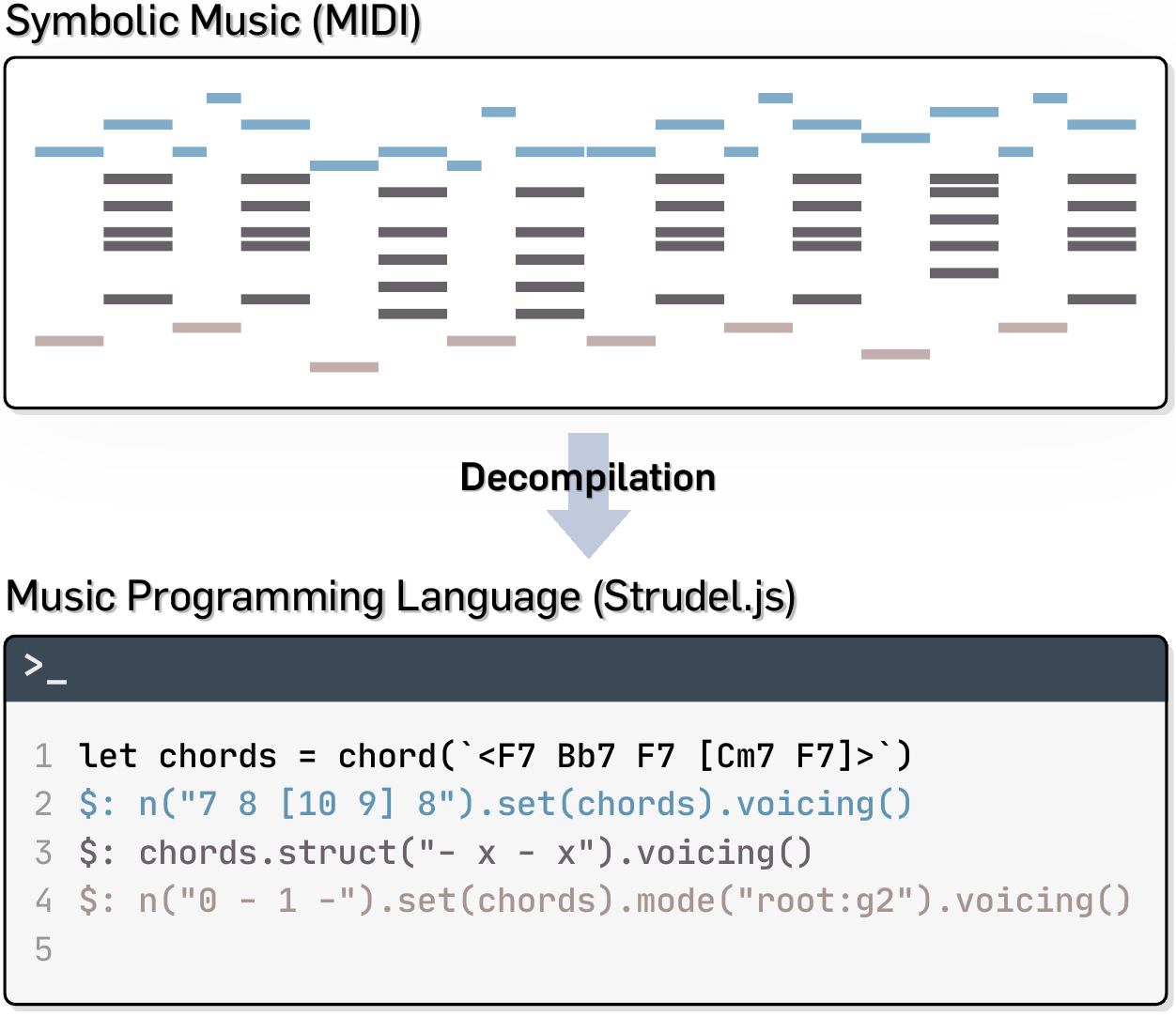}
    \caption{\textbf{Symbolic Music Decompilation.} 
    In this task, the goal is to generate a \textit{faithful} and \textit{readable} music program from symbolic music input.
    We instantiate this task 
    as MIDI-to-\Strudel{} decompilation; 
    \Strudel{} is a domain-specific music programming language for algorithmic composition and live coding. 
    The recovered program should reproduce the input when executed while exposing readable and editable musical structure.
    }
    \label{fig:music_decompilation}
    \vspace{-1em}
\end{wrapfigure}

%% file: figures/fig_system.tex
\begin{figure*}[t!]
    \centering
    \includegraphics[width=\linewidth]{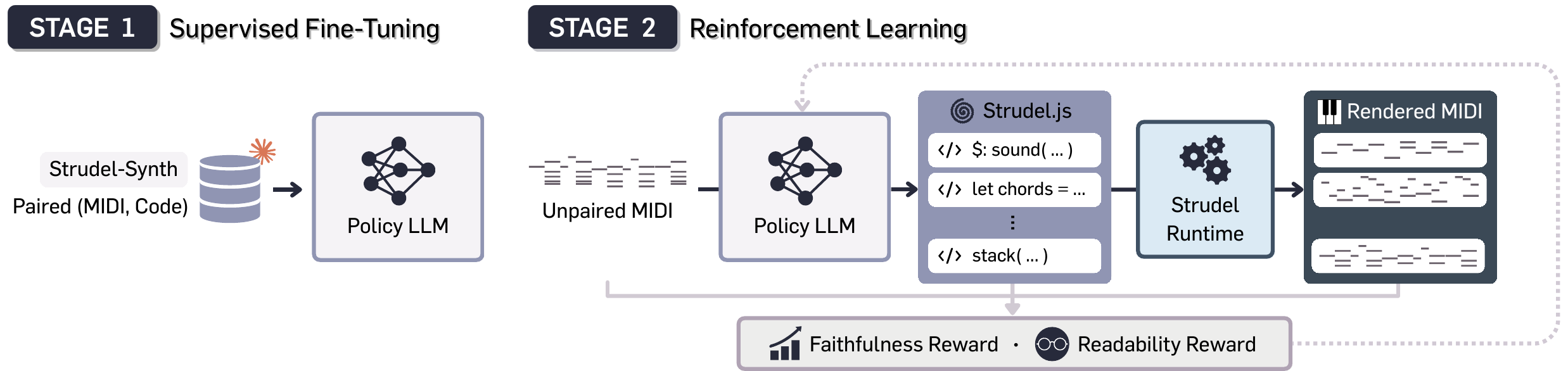}
    \caption{
    \textbf{\system{} Overview.}
    We present a framework for post-training LLMs to 
    decompile symbolic music (MIDI) into a music programming language (\Strudel{}). 
    \system{} consists of two stages: 
    (i) supervised fine-tuning equips the model with the ability to write valid Strudel code; 
    (ii) reinforcement learning optimizes the decompilation objective directly, rewarding sampled programs for both \emph{faithfulness} (rendered MIDI matches the input) and \emph{readability} (concise, editable code).
    The paired examples used for SFT are provided by \dataset{}, our synthetic corpus of executable \Strudel{} programs and rendered MIDI.
    Since the reward depends only on the input MIDI and executed program, RL needs no reference code and trains on any unpaired MIDI at scale.
    }
    \vspace{-1em}
    \label{fig:main}
\end{figure*}

%% file: sections/3_method.tex
\section{\system{}}
\label{sec:method}

We approach symbolic music decompilation 
as program synthesis with execution feedback,
following recent decompilation work that optimizes programs against a 
verifiable renderer~\citep{kolodiazhnyi2026cadrille, juan2025rlrfsvg}. 
Given an input MIDI segment $x$, 
the goal is to
model $p(y \mid x)$ 
where $y$ is a corresponding Strudel program.
Specifically, 
for the \Strudel{} runtime $\mathcal{C}$ 
and a program $\hat{y} \sim p(y \mid x)$, 
our goal is to improve faithfulness of the MIDI output from Strudel execution ($\mathcal{C}(\hat{y}) \approx x$), while ensuring $\hat{y}$ is human-readable. 
We parameterize a conditional language model
policy
\begin{equation}
    \label{eq:policy}
    \pi_\theta(y \mid x)
=
\texttt{LLM}_\theta(\texttt{MidiToText}(x)),
\end{equation}
where \texttt{MidiToText} 
is a helper function that 
converts the MIDI input into a text representation
compatible with a general LLM (see Appendix~\ref{app:midi_repr}). 
Because many \Strudel{} programs 
can render to the same or similar MIDI output (see Appendix~\ref{app:strudel:examples} for examples), 
decompilation quality is
defined by both rendered-MIDI faithfulness and code-level readability.

The remainder of this section 
first introduces our two-stage training pipeline
combining SFT and RL (Section~\ref{sec:method:training}) and details the composite
reward used in the RL stage (Section~\ref{sec:method:reward}). Finally, we describe
\dataset{}, the synthetic paired (MIDI, \Strudel{}) corpus we construct
(Section~\ref{sec:method:dataset}).

\newcommand{\dmidi}{\mathcal{D}_{\text{M}}}
\newcommand{\dstrudel}{\mathcal{D}_{\text{S}}}

\subsection{Two-stage training}
\label{sec:method:training}

Directly applying 
RL to this task from a 
general-purpose open-weight LLM is difficult: 
unlike common LLM post-training domains such as math or conventional code generation, 
MIDI-to-\Strudel{} decompilation 
requires grounding a dense symbolic music input 
while generating a low-resource DSL 
that appears rarely in
pretraining data.
As a result, early rollouts from the unadapted model 
seldom produce valid \Strudel{} programs, 
making reward signals sparse and unstable. 
We therefore use a two-stage pipeline: supervised fine-tuning first gives the model
a prior over executable \Strudel{} code, and reinforcement learning then
optimizes the decompilation objective.
 
\paragraph{Stage 1: Supervised fine-tuning (SFT).}
We first adapt the base LLM $\pi_\theta$ 
to the MIDI-to-Strudel
decompilation task 
using the paired dataset $\dstrudel$. 
Each example
$(x^*, y^*) \in \dstrudel$ consists of 
a Strudel program $y^*$ and its executed MIDI representation $x^* = \mathcal{C}(y^*)$, 
where $\mathcal{C}$ denotes the
Strudel runtime.\footnote{We use $y^*$ to indicate that, during SFT,
the target program corresponding to $x^* = \mathcal{C}(y^*)$ is known, in
contrast to a generic MIDI input $x$ at test time or during RL where the target program is unknown.}
Given the MIDI input $x^*$, the model is trained to generate 
the corresponding program $y^*$ with the standard next-token prediction objective:

\begin{equation}\label{eq:sft}
    \mathcal{L}_{\text{SFT}}(\theta) =
    -\, \mathbb{E}_{y^* \sim \dstrudel}    \!\left[\,\sum_{t=1}^{|y^*|} \log \pi_\theta(y^*_t \mid y^*_{<t},\, x^*)\right].
\end{equation}

Without this stage, 
RL collapses onto a degenerate solution where the model learns to emit short, literal note lists that compile but capture almost none of the input music, and reward plateaus early at a low value. 
SFT instead provides a warm start
that places the policy in a region on program space where rollouts express nontrivial musical structure, so that the gradient signal from the faithfulness and readability terms is meaningful (Section~\ref{sec:exp:ablation}).

\paragraph{Stage 2: Reinforcement learning.}
After SFT, we further train the model to optimize the 
decompilation objective:
a generated program should 
faithfully reproduce the input MIDI 
when executed
while remaining readable. 
We use the SFT model
$\pi_{\theta}$ as 
the initial policy and optimize it by maximizing the expected reward
\begin{equation}
    \label{eq:objective}
    \mathcal{J}(\theta)
    =
    \mathbb{E}_{x \sim \dmidi,\; \hat{y} \sim \pi_\theta(\cdot \mid x)}
    \left[ R(\hat{y}, x) \right],
\end{equation}

where $\dmidi$ is a dataset of unpaired MIDI and 
$R$ is a composite reward combining 
decompilation faithfulness 
and code readability (see Section~\ref{sec:method:reward}).
Because $R$ depends only on the input MIDI 
and the executed program, 
this stage requires no reference \Strudel{} code 
and can train on any unpaired MIDI at scale.

To optimize Equation~\ref{eq:objective}, 
we adopt Group reward-Decoupled normalization Policy Optimization (GDPO;~\citealt{gdpo}),
a GRPO-style policy optimization method~\citep{guo2025deepseek} for multi-reward objectives.
For each input MIDI $x$, 
the policy samples a group of $G>0$ candidate programs, 
executes them with the \Strudel{} runtime,
and evaluates each program with $K$ reward components $r_1,\ldots,r_K$.
In our setting, $K=2$, 
corresponding to faithfulness and readability.
A standard GRPO-style update 
first scalarizes these components as
$r^{(g)}=\sum_k w_k r_k^{(g)}$ 
where $w_k$ are reward weights, 
and then normalizes the scalar rewards across the group; 
GDPO instead 
constructs a group-relative advantage for each reward component 
and combines the normalized advantages only afterward.

Concretely, let $r_k^{(g)}$ denote the $k$-th reward component of the $g$-th sampled program, 
where $k \in \{1,\ldots,K\}$ and $g \in \{1,\ldots,G\}$.
For each component, GDPO computes
\begin{equation}
\label{eq:gdpo}
A_k^{(g)}
=
\frac{r_k^{(g)}-\mu_k}{\sigma_k+\epsilon},
\quad
\mu_k=\frac{1}{G}\sum_{g'=1}^{G}r_k^{(g')},
\quad
\sigma_k=\mathrm{std}\!\left(\{r_k^{(g')}\}_{g'=1}^{G}\right).
\end{equation}
The final advantage is 
$
A^{(g)}=\sum_{k=1}^{K} w_k A_k^{(g)},
$
which is used in the standard clipped group-relative policy objective to update $\pi_\theta$.
Full optimization details are provided in Appendix~\ref{app:grpogdpo}.

\subsection{Reward}
\label{sec:method:reward}

We design a reward function 
$R(\hat{y},x)$
that evaluates a generated
\Strudel{} program $\hat{y}$ for an input MIDI segment $x$ along two axes:
\emph{faithfulness}, whether the rendered MIDI $\hat{x} = \mathcal{C}(\hat{y})$ matches the input $x$, and
\emph{readability}, whether the generated program $\hat{y}$ is easy to understand and edit.

\paragraph{Faithfulness.}
The faithfulness reward 
$R_{\text{faith}}(\hat{x},x)\in[0,1]$
measures how accurately the rendered MIDI 
$\hat{x}=\mathcal{C}(\hat{y})$ reproduces the input MIDI $x$. 
We adopt the instrument-aware onset metric from 
the multi-track automatic music transcription literature~\citep{mt3, lu2023multitrack, yourmt3, andrew2018evaluating}: 
$R_{\text{faith}}$
is the multi-instrument onset F1 
between $\hat{x}$ and $x$,
where a predicted note is counted as
correct only if it 
(i) matches the pitch of a reference note, 
(ii) has an onset
within $\pm 50$\,ms of that reference onset, 
and (iii) is assigned the same
General MIDI program number as the reference note.
The optimal pairing between
reference and predicted notes is obtained by bipartite matching, and 
$R_{\mathrm{faith}}$ is the resulting F1, computed with the standard
\texttt{mir\_eval} implementation~\citep{mireval}.

\paragraph{Readability.}

The readability reward
$R_{\text{read}}(\hat{y}) \in [0, 1]$ measures whether the generated program $\hat{y}$ is
easy to read and edit.
This term is essential because 
optimizing $R_{\text{faith}}$ alone 
admits a degenerate solution: 
the model can learn a literal, note-by-note transliteration that
renders correctly but captures none of the repeated patterns or musical structure that make
\Strudel{} concise and editable.

Following the rubric-as-rewards paradigm~\citep{viswanathan2025checklists, gunjal2026rubrics}, 
we therefore score readability with a fixed set of $J=12$ rubric items, each a binary yes/no check evaluated by an LLM judge.
The items cover 
(i) general code readability such as formatting and structure, 
adapted from established code-readability metrics~\citep{bnw, scalabrino2018readability};
and (ii) \Strudel{}-specific criteria targeting 
concision and musical abstraction, 
such as
using pattern operators and chord symbols
rather than literal pitches (full rubric in Appendix~\ref{app:rubrics}).
To turn these checks into a scalar reward, we 
let $c_j : \hat{y} \mapsto \{0,1\}$ 
indicate whether a sampled program $\hat{y}$ satisfies rubric item
$j$ 
and aggregate with uniform weights:
\begin{equation}
\label{eq:read}
R_{\text{read}}(\hat{y})
=
\frac{1}{J}\sum_{j=1}^{J} c_j(\hat{y}).
\end{equation}

\paragraph{Final reward.}
Both rewards presume an executable program:
a non-compiling output yields no rendered MIDI
and exposes no executable structure for the readability judge to credit. 
We therefore gate each reward on successful compilation, assigning a penalty $-\rho$ ($\rho>0$) when the \Strudel{} runtime $\mathcal{C}$ fails to render $\hat{y}$:
\begin{equation}
\label{eq:reward}
\hat{R}_{m}
=
\begin{cases}
R_{m}, & \text{if } \mathcal{C}(\hat{y}) \text{ succeeds}, \\
-\rho, & \text{otherwise},
\end{cases}
\qquad
m \in \{\mathrm{faith}, \mathrm{read}\}.
\end{equation}

The gated rewards \mbox{($\hat{R}_{\mathrm{faith}}$, $\hat{R}_{\mathrm{read}}$)} 
form the two components \mbox{$(r_1, r_2)$} 
that GDPO normalizes separately before
combining (Section~\ref{sec:method:training}).

\subsection{Dataset: \dataset{}}
\label{sec:method:dataset}

Our SFT warm start requires
a paired \mbox{(MIDI, code)} corpus,
which the public web does not supply at scale:
crawling public \Strudel{} programs
yields only $688$ non-trivial human-authored examples.
Following prior work on synthetic data generation
~\citep{doh2025talkplay, doh2026llmfxtools, mukherjee2023orca},
we construct \dataset{} by distilling from a frontier LLM (\claude{})
to write \Strudel{} programs and executing each through the 
\Strudel{} runtime $\mathcal{C}$ to obtain its rendered MIDI.

\paragraph{Data generation pipeline.}
\input{figures/fig_dataset_pipeline}
Our pipeline produces \mbox{(MIDI, \Strudel{} code)} pairs in three stages
(Figure~\ref{fig:dataset_pipeline}).
We begin by generating \Strudel{} programs from an LLM, 
conditioned on a musical seed $s_{\text{music}}$ 
and a code style seed $s_{\text{code}}$ 
under a fixed task prompt. 
In the second stage, we clean the generated programs, 
as LLM-written programs
frequently include dead voices: 
layers that compile but emit no MIDI events, typically from hallucinated instrument names or malformed operator chains. 
We first apply deterministic regex rewrites that correct these recurring mistakes, 
then use the Acorn JavaScript parser to split each program into its
independent voices, compile each voice in isolation, and prune the silent ones together with the variable declarations they alone reference. 
A program is
retained only if at least one voice survives.
Lastly, 
we render each cleaned program to MIDI with the runtime $\mathcal{C}$.
Full details of the pipeline are given in Appendix~\ref{app:dataset:pipeline}.

\paragraph{Conditioning inputs and tools.}
To ensure diversity in \dataset{},
we independently vary two conditioning seeds during generation: 
a musical seed $s_{\text{music}}$ and 
a code-style seed $s_{\text{code}}$, 
which respectively specify the musical content of the piece and 
how it should be written. 
Specifically, 
$s_\text{music}$ samples musical attributes---genre, mood, key, tempo, meter, chords, and instrumentation---from existing music metadata~\citep{mcdonald_everynoise1d, midicaps, theorytab}, 
and we vary how many of these attributes are specified during generation. 
The code-style seed 
$s_\text{code}$
selects among different program idioms that 
realize the same musical target,
ranging from 
mini-notation (\texttt{\$}) based patterns 
to multi-section programs using \texttt{arrange} functions\footnote{\prettyurl{https://strudel.cc/learn/factories}}. 
We render each generated program to MIDI with a headless \Strudel{} runtime $\mathcal{C}$ and use the same deterministic renderer to score policy rollouts during RL (Section~\ref{sec:method}). 
Further details on the conditioning inputs and runtime are provided in Appendix~\ref{app:dataset:conditioning} and~\ref{app:strudel:runtime}, respectively.

\paragraph{Statistics.}
\input{tables/dataset_stats}
Table~\ref{tab:dataset_stats} presents the statistics of \dataset{}.
It comprises $21{,}174$ \mbox{(MIDI, \Strudel{})} pairs generated from \claude{} with diverse
conditioning seeds,
roughly $30\times$ the $688$ non-trivial programs found from public \Strudel{} crawl.
The generated programs cover 
$122$ of the $128$ General MIDI instruments
and render to short musical fragments, 
averaging $24$ seconds in duration 
(from $2$s to $90$s).

%% file: figures/fig_dataset_pipeline.tex
\begin{wrapfigure}{r}{0.5\textwidth}
\vspace{-0.2in}
    \includegraphics[width=\linewidth]{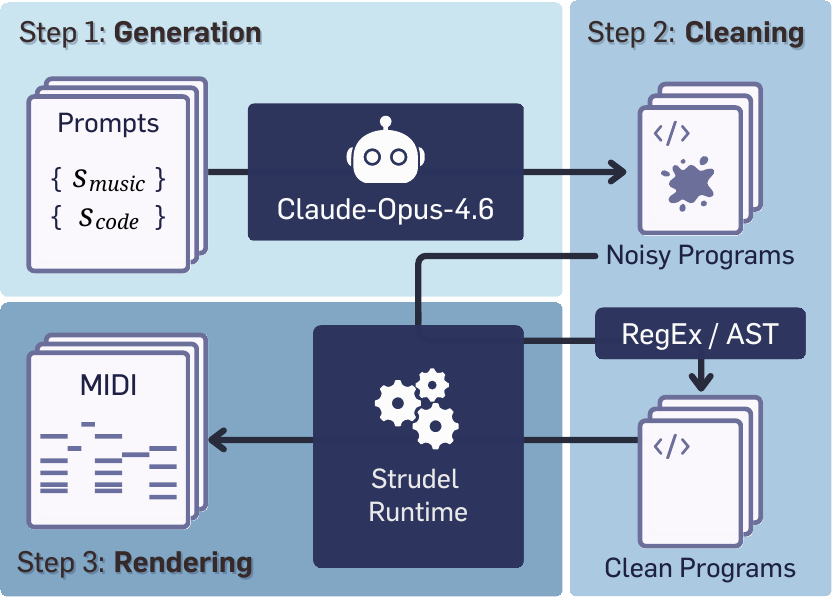}
    \caption{\textbf{\dataset{} generation pipeline.} 
    Each \mbox{(MIDI, \Strudel{} code)} pair is produced in three stages:
    (i) prompting an LLM with a musical and a code style seed, 
    (ii) cleaning the generated program, and 
    (iii) rendering it to MIDI through the \Strudel{} runtime. 
    }
    \label{fig:dataset_pipeline}
\end{wrapfigure}

%% file: tables/dataset_stats.tex
\begin{wraptable}{r}{0.52\linewidth}
\vspace{-0.95em}
\centering
\footnotesize
\caption{
    \textbf{Statistics of \dataset{} dataset.} 
}
\setlength{\tabcolsep}{4.5pt}
\renewcommand{\arraystretch}{1.18}
\begin{tabular}{@{}l c c@{}}
\toprule
\textbf{Metric} & \textbf{Train Split} & \textbf{Test Split} \\
\midrule
\# of (MIDI, Code) Pairs
& 20,152
& 1,022
\\
\# of Instruments
& 122
& 98
\\
Avg. \# Insts / Song
& 3.73{\scriptsize$\pm$1.22}
& 3.68{\scriptsize$\pm$1.26}
\\
Avg. Lines of Code
& 52.20{\scriptsize$\pm$20.88}
& 52.46{\scriptsize$\pm$21.70}
\\
Avg. Duration (s)
& 24.17{\scriptsize$\pm$19.03}
& 24.10{\scriptsize$\pm$18.70}
\\
\bottomrule
\end{tabular}
\label{tab:dataset_stats}
\vspace{-0.4em}
\vspace{-0.9em}
\end{wraptable}

%% file: sections/4_experiments.tex
\section{Experiments}
\label{sec:exp}

We evaluate the performance of \system{} 
and the effect of its proposed components 
on both synthetic and real-world MIDI data. 
Our experiments address the following questions:

\begin{itemize}
[leftmargin=*,itemsep=0mm]
    \item Can \system{} recover faithful, readable \Strudel{} programs from MIDI? (Section~\ref{sec:exp:main})
    \item How do \system{}'s components affect decompilation quality? (Section~\ref{sec:exp:ablation})
    \item Can \system{} generalize across diverse MIDI datasets? (Section~\ref{sec:exp:generalization})
    \item What downstream applications does music decompilation enable? (Section~\ref{sec:exp:application})
\end{itemize}

\subsection{Setup}
\label{sec:exp:setup}

\paragraph{Metrics.}
We evaluate \system{} 
along three axes: 
faithfulness, readability, and diversity.
For faithfulness, 
we report compile rate (Comp.)
together with transcription metrics 
adapted from automatic music transcription~\citep{mt3, maman2022unaligned, yourmt3}: 
Onset F1, Frame F1, and Multi-Instrument Onset F1 (Inst F1).
Readability is measured by 
the proposed rubric score (Rubric) 
and average rank from a list-wise LLM reranker (Rank;~\citealp{tang2024found}).
Diversity is measured by 
the average pairwise CodeBLEU 
self-similarity
(SelfCB;~\citealp{ren2020codebleu}) 
among compiled outputs from the same method. 
Further details are provided in Appendix~\ref{app:eval:metrics}.

\paragraph{Datasets.}
Our main experiments use two corpora: \dataset{} (synthetic MIDI--Strudel pairs) and LMD (real-world MIDI; \citealp{lakhmidi}).
For LMD, we use short fragments (${<}30$\,s; ${\sim}9$K train / ${\sim}1$K test).
To test generalization beyond these primary datasets, we further evaluate on four held-out datasets spanning different MIDI distributions: 
longer LMD ($30$--$60$\,s), 
GigaMIDI~\citep{lee2025gigamidi}, 
NES-MDB~\citep{donahue2018nesmdb}, and
Nottingham~\citep{foxley_nottingham}. 
Except for longer LMD, all held-out corpora use short fragments 
(${<}30$\,s); see Appendix~\ref{app:eval:dataset} for details.

\paragraph{Baselines.}
Since MIDI-to-Strudel decompilation 
has no direct prior baseline, 
we compare against two classes of baselines.
First, we prompt frontier LLMs to 
generate Strudel code from the MIDI input: \claude~\citep{anthropic2026opus46}, \gemini~\citep{google2026gemini35flash}, and 
\gpt~\citep{openai2026gpt55}. 
Second, we include a heuristic MIDI-to-Strudel converter~\citep{midi_to_strudel}, 
which directly transliterates MIDI events into Strudel code. 
Baseline prompts and decoding settings are described in Appendix~\ref{app:eval:baseline}.

\paragraph{Implementation details.}
All experiments use 
open-weight models from the Qwen3 family~\citep{qwen3technicalreport},
specifically 4B (\texttt{Qwen3-4B-Instruct-2507}) and 8B (\texttt{Qwen3-8B}) models.
We split the \dataset{}'s training set into two disjoint halves of 
approximately 10K samples each.
The first half is used for SFT with paired \mbox{(MIDI, Strudel)} examples;
the second half is used for RL, where only the MIDI input is provided.
For SFT, we fine-tune each model for one epoch with LoRA~\citep{hu2022lora}.
For RL, we initialize from the SFT checkpoint and train on 
a mixture of \dataset{} MIDI and short-fragment (<30s) LMD MIDI.
Unless otherwise specified, 
we set the reward weights to $(w_1, w_2) = (1.0, 0.5)$ 
for faithfulness and readability.
Additional hyperparameters and compute details
are provided in Appendix~\ref{app:eval:hyperparams}.

\input{tables/main_table}

\subsection{Main Results}
\label{sec:exp:main}

Table~\ref{tab:main} summarizes the main quantitative results. 
Overall, \system{} 
achieves a more favorable faithfulness--readability tradeoff 
than any baseline: 
it is substantially more faithful than frontier LLMs 
while producing far more readable code than the heuristic converter.

\paragraph{\system{} is more faithful than frontier LLMs.}
Frontier LLMs (\claude{}, \gemini{}, \gpt{}) 
can generate executable \Strudel{} code, 
yet their renders often diverge from the input MIDI. 
This gap widens on LMD: 
compared with the best frontier LLM, 
\system{} (8B) improves Onset F1 
by $+0.16$ on \dataset{} 
and by $+0.32$ on LMD, suggesting
that real-world MIDI requires stronger input grounding than general-purpose LLMs provide.

\paragraph{\system{} avoids the unreadable outputs of heuristic conversion.} 
The heuristic converter is nearly perfectly faithful by construction,
but its note-by-note transliteration yields the lowest readability of all methods 
(rubric score 0.05 on \dataset{} and 0.09 on LMD; worst average rank on both).
\system{} instead attains 
rubric scores an order of magnitude higher (0.61--0.74) and 
ranks comparably to frontier LLMs. 

\paragraph{\system{} avoids diversity collapse.} 
Compared with SFT alone, 
applying RL increases SelfCB: 
for the 8B model, 
SelfCB rises by $+0.11$ on \dataset{} and $+0.10$ on LMD. 
This is consistent with prior work showing that reward optimization can reduce output diversity~\citep{wang2026optimizing, west2025base, noveltybench}. 
Importantly, its SelfCB remains well below that of the heuristic converter 
($-0.17$ on \dataset{} and $-0.30$ on LMD) and comparable to frontier LLMs, 
showing that \system{} does not collapse to a fixed transliteration template.

\paragraph{Execution-based RL improves real-world decompilation.} 
SFT alone transfers poorly to real-world MIDI:
for the 8B model, SFT improves
Onset F1 by $+0.41$ on \dataset{} but only by $+0.06$ on LMD.
Execution-based RL shows the complementary effect, 
giving a larger gain on LMD 
than on \dataset{} ($+0.50$ vs. $+0.27$ Onset F1).
This improvement requires 
no reference \Strudel{} programs during RL, 
showing that \system{} can bootstrap from synthetic paired data 
and then adapt to real-world MIDI 
through label-free execution feedback.

\subsection{Ablation studies}
\label{sec:exp:ablation}

\paragraph{Effect of readability weight.}
We vary the readability weight $w_2$ while keeping the faithfulness weight fixed
at $w_1=1.0$.
Figure~\ref{fig:ablation_weights} reveals a clear faithfulness--readability tradeoff:
with $w_2=0$, the model produces programs that better match the input MIDI 
but are often verbose and close to note-by-note transcriptions.
Adding a readability reward shifts the model toward more abstract and editable code, yet sacrificing faithfulness. 
We set $w_2=0.5$ for our main experiments, as it improves readability
substantially without the large faithfulness drop observed at higher weights.

\paragraph{Effect of SFT initialization.}
We compare RL initialized from the SFT checkpoint 
against RL initialized directly from
the unadapted \texttt{Qwen3-8B} model. 
As shown in Figure~\ref{fig:ablation_sft}, 
RL without SFT plateaus early: 
the policy collapses to short, literal note lists 
that compile but capture little of the input, 
and readability does not improve. 
SFT places the policy into a region where sampled programs compile and express nontrivial musical structure, making execution-based RL feasible. 

\input{figures/fig_ablations}

\subsection{Generalization Across Datasets}
\label{sec:exp:generalization}

We test whether \system{} 
generalizes beyond the training distributions 
using the 8B model on four datasets: 
longer LMD ($30$--$60$s), 
GigaMIDI, 
NES-MDB, 
and Nottingham, 
covering shifts in 
duration, 
corpus source,
genre, 
and instrumentation.

\input{tables/generalization_table}
Table~\ref{tab:generalization} shows that 
\system{} outperforms GPT-5.5 
across all evaluated distributions. 
The model transfers well to Nottingham and GigaMIDI, 
suggesting that it can recover compact programmatic structure beyond
the synthetic Strudel distribution. 
Performance is lower on longer LMD excerpts and NES-MDB, 
indicating that robustness still varies across distributions,
especially when inputs become longer or differ more substantially from the
training data.
At the same time, 
these limitations show a practical advantage of our pipeline: 
because the RL stage requires only unpaired MIDI, 
such held-out real-world corpora can be incorporated 
into post-training without reference Strudel code.

\subsection{Applications of \system{}}
\label{sec:exp:application}

We discuss how \system{} can support downstream music workflows 
by turning music into executable, editable code. 
We also provide proof-of-concept examples on our project page.

\paragraph{Music decompilation (audio-to-code).}
Combined with automatic music transcription~\citep{mt3, yourmt3, shin2026music}, \system{} provides a path toward 
audio-to-code decompilation: 
audio is first transcribed to MIDI, and the resulting MIDI is then decompiled into a \Strudel{} program.
Since MIDI only partially captures acoustic attributes such as timbre, dynamics, and audio effects, extending music decompilation 
to audio would require modeling these attributes, 
potentially using audio production modeling methods~\citep{Steinmetz2024STITO, comunita2025differentiable}
or audio language models with music understanding~\citep{Qwen2.5-Omni, musicflamingo}.

\paragraph{Music creation as code editing.} 
\system{} provides a programmatic editing interface for MIDI 
by converting sequential note events into \Strudel{} programs 
that expose musical structures such as repetition, layering, harmony, and rhythm.
These programs could be used 
in live coding settings or alongside DAW workflows, 
allowing users to edit higher-level 
musical structures 
such as chord progressions or scales in code 
rather than adjusting individual notes manually.
Moreover, 
in creative settings where exploring multiple alternatives matters more than a single fixed output~\citep{resnick2005design, kim2025amuse},
the faithfulness--readability tradeoff (Section~\ref{sec:exp:ablation}) could be a 
controllable feature: 
users can dial the system toward faithful transcriptions for precise editing, or toward abstraction for creative exploration.

\paragraph{Music understanding through code.}
Recovering code from MIDI can make musical structure easier to inspect and reason about.
Unlike a low-level 
MIDI sequence, 
a \Strudel{} program with explicit musical structure 
could serve as an 
intermediate representation for LLM-based music understanding~\citep{gardner2023llark, musicflamingo}, 
allowing models to reason over compact, interpretable programs rather than long sequences of note events.
Such programs could also benefit human users: decompiled code could provide interactive material for music education~\citep{chris2022sonic}, 
and its explicit repetition could aid analysis tasks such as motif discovery~\citep{boot2016evaluating}.

%% file: tables/main_table.tex
\begin{table*}[t]
\setlength{\tabcolsep}{4pt} 
\centering
\small
\caption{
\textbf{Main results across synthetic and real-world MIDI.} 
We report results on held-out splits of \dataset{} and LMD, matching the distributions used for training.
\system{} improves the faithfulness--readability tradeoff for symbolic music decompilation:
it outperforms frontier LLMs in execution faithfulness
while producing substantially more readable and diverse code than the heuristic converter. $\downarrow$ and $\uparrow$ indicate 
whether lower or higher values are better, respectively. 
}
\begin{tabular}{clccccccc}
\toprule
&& \multicolumn{4}{c}{\textbf{Faithfulness}} 
& \multicolumn{2}{c}{\textbf{Readability}}  & \textbf{Diversity} \\ 
\cmidrule(lr){3-6} \cmidrule(lr){7-8} \cmidrule(lr){9-9}
\textbf{Dataset} & \textbf{Method} 
& $\uparrow$Comp. & $\uparrow$Onset & $\uparrow$Frame & $\uparrow$Inst
& $\uparrow$Rubric & $\downarrow$Rank & $\downarrow$SelfCB \\
\midrule

\multirow{12}{*}{\textbf{\makecell{\textsc{Strudel}\\-\textsc{Synth}}}}
& Heuristic & 1.00 & 0.99 & 0.83 & 0.99 & 0.05 & 7.89 & 0.72 \\ 
%
& \claude{} & 0.73 & 0.50 & 0.59 & 0.42 & 0.42 & 4.41 & 0.46 \\ 
& \gemini{} & 0.57 & 0.50 & 0.59 & 0.42 & 0.45 & 3.75 & 0.46 \\ 
& \gpt{} & 0.80 & 0.57 & 0.63 & 0.50 & 0.39 & 2.76 & 0.47 \\ 
\cmidrule(){2-9}
& Qwen3-4B & 0.06 & 0.03 & 0.08 & 0.01 & 0.37 & -- & -- \\ 
& \,\,+ SFT & 0.58 & 0.41 & 0.46 & 0.39 & 0.58 & 4.76 & 0.42 \\ 
& \cellcolor{ypaleblue}\,\,\textbf{+ SFT+RL (ours)} & 
\cellcolor{ypaleblue}0.99 & 
\cellcolor{ypaleblue}0.71 & 
\cellcolor{ypaleblue}0.67 & 
\cellcolor{ypaleblue}0.70 & 
\cellcolor{ypaleblue}0.73 & 
\cellcolor{ypaleblue}3.65 & 
\cellcolor{ypaleblue}0.56 \\
\cmidrule(){2-9}
& Qwen3-8B & 0.19 & 0.05 & 0.10 & 0.02 & 0.33 & -- & -- \\ 
& \,\,+ SFT & 0.62 & 0.46 & 0.46 & 0.41 & 0.58 & 4.96 & 0.44\\ 
& \cellcolor{ypaleblue}\,\,\textbf{+ SFT+RL (ours)} & 
\cellcolor{ypaleblue}0.99 & 
\cellcolor{ypaleblue}0.73 & 
\cellcolor{ypaleblue}0.69 & 
\cellcolor{ypaleblue}0.72 &
\cellcolor{ypaleblue}0.74 & 
\cellcolor{ypaleblue}3.83 & 
\cellcolor{ypaleblue}0.55\\

\midrule

\multirow{12}{*}{\textbf{LMD}}
& Heuristic & 1.00 & 0.95 & 0.67 & 0.89 & 0.09 & 7.62 & 0.77 \\ 
& \claude{} & 0.75 & 0.28 & 0.33 & 0.21 & 0.36 & 4.47 & 0.60 \\ 
& \gemini{} & 0.63 & 0.22 & 0.30 & 0.17 & 0.34 & 4.37 & 0.51 \\ 
& \gpt{} & 0.82  & 0.27 & 0.31 & 0.23 & 0.29 & 4.27 & 0.56 \\ 
\cmidrule(){2-9}
& Qwen3-4B & 0.17 & 0.04 & 0.12 & 0.03 & 0.35 & -- & -- \\ 
& \,\,+ SFT & 0.45 & 0.09 & 0.13 & 0.07 & 0.55 & 4.01 & 0.38 \\ 
& \cellcolor{ypaleblue}\,\,\textbf{+ SFT+RL (ours)} & 
\cellcolor{ypaleblue}0.99 & 
\cellcolor{ypaleblue}0.54 & 
\cellcolor{ypaleblue}0.49 & 
\cellcolor{ypaleblue}0.52 & 
\cellcolor{ypaleblue}0.70 & 
\cellcolor{ypaleblue}3.13 & 
\cellcolor{ypaleblue}0.53 \\
\cmidrule(){2-9}
& Qwen3-8B & 0.18 & 0.04 & 0.11 & 0.02  & 0.28 & -- & -- \\
& \,\,+ SFT & 0.44 & 0.10 & 0.14 & 0.09 & 0.54 & 4.01 & 0.37 \\
& \cellcolor{ypaleblue}\,\,+ \textbf{SFT+RL (ours)} & 
\cellcolor{ypaleblue}0.99 & 
\cellcolor{ypaleblue}0.60 & 
\cellcolor{ypaleblue}0.53 & 
\cellcolor{ypaleblue}0.58 &
\cellcolor{ypaleblue}0.61 & 
\cellcolor{ypaleblue}4.12 & 
\cellcolor{ypaleblue}0.47 \\

\bottomrule
\end{tabular}
\label{tab:main}
\end{table*}

%% file: figures/fig_ablations.tex
\begin{figure*}[t!]
\vspace{-0.1in}
\centering\small

\begin{minipage}[t]{0.33\textwidth}
    \centering
    \includegraphics[width=0.985\linewidth]{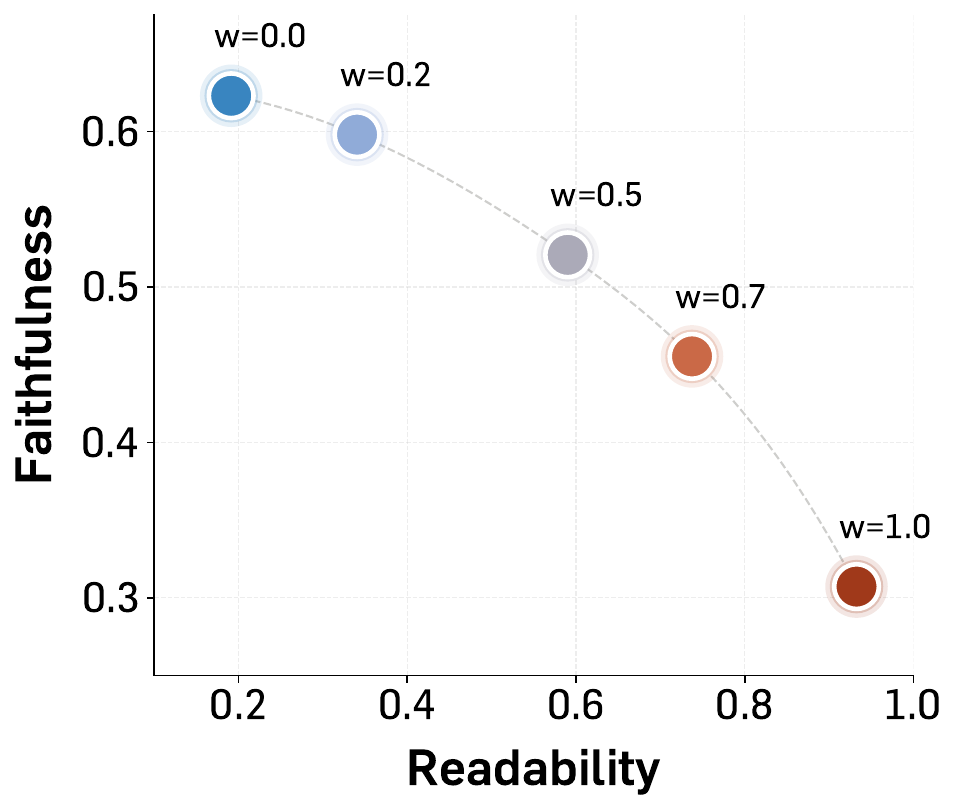}
    \caption{
    \textbf{Effect of readability weight ($w_2$).} 
    We vary $w_2$ for Qwen3-8B with fixed $w_1=1.0$.
    Larger $w_2$ improves readability at the cost of faithfulness.
    }
    \label{fig:ablation_weights}
\end{minipage}%
\hfill
\begin{minipage}[t]{0.65\textwidth}
    \centering
    \begin{subfigure}[t]{0.5\linewidth}
        \centering
        \includegraphics[width=\linewidth]{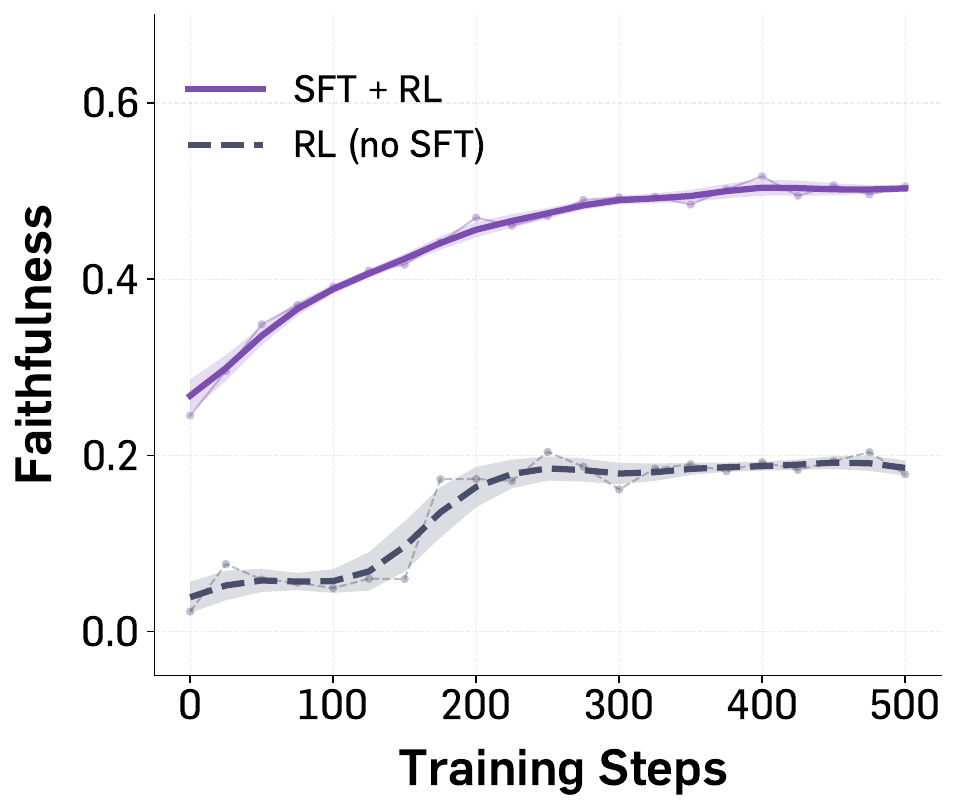}
    \end{subfigure}%
    \hfill
    \begin{subfigure}[t]{0.5\linewidth}
        \centering
        \includegraphics[width=\linewidth]{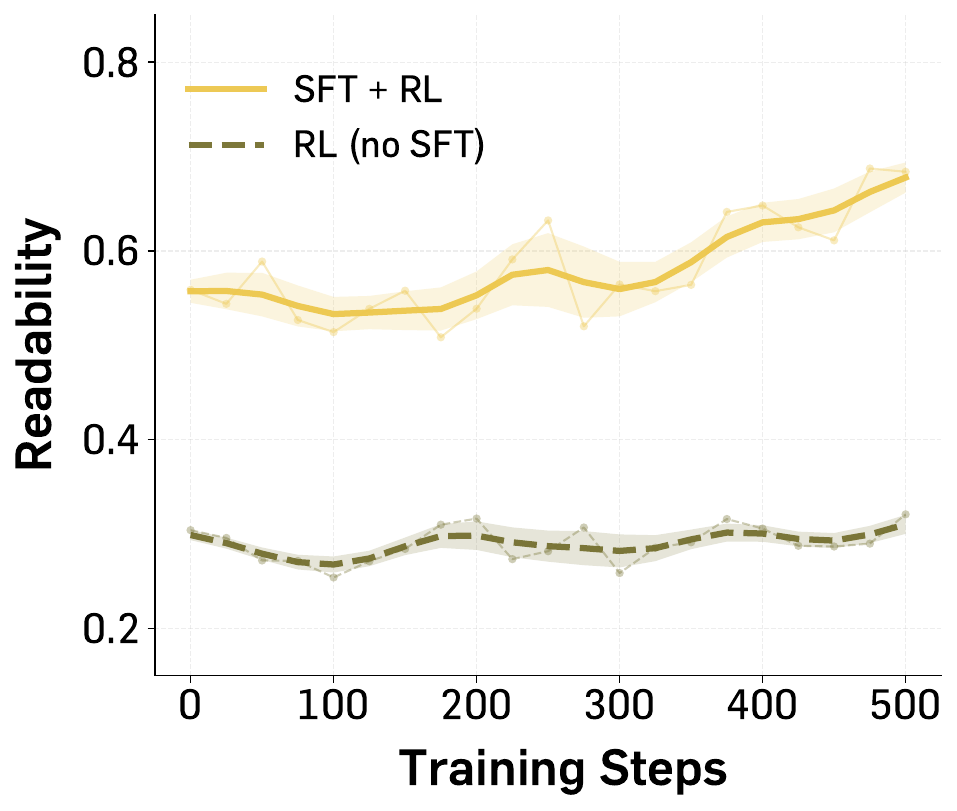}
    \end{subfigure}
    \caption{\textbf{Effect of SFT initialization.} 
    We show faithfulness (left) and readability (right) rewards 
    during RL training on Qwen3-8B, 
    with and without SFT initialization. 
    Without SFT, RL plateaus early at a degenerate low-reward solution; 
    SFT initialization provides a warm start that enables both rewards to improve steadily.
    }
    \label{fig:ablation_sft}
\end{minipage}

\end{figure*}

%% file: tables/generalization_table.tex
\begin{wraptable}{r}{0.52\linewidth}
\vspace{-0.95em}
\centering
\footnotesize
\caption{
    \textbf{Generalization across datasets.}
    We report Best@5 Onset F1 and Rubric, selected by Inst F1; \system{} consistently outperforms \gpt{}.
}
\setlength{\tabcolsep}{4.5pt}
\renewcommand{\arraystretch}{1.18}
\begin{tabular}{@{}lcccc@{}}
\toprule
& 
\multicolumn{2}{c}{\textbf{GPT-5.5}} &
\multicolumn{2}{c}{\textbf{\system{}}}
\\
\cmidrule(lr){2-3} \cmidrule(lr){4-5}
\textbf{Dataset} & $\uparrow$Onset & $\uparrow$ Rubric & $\uparrow$Onset & $\uparrow$ Rubric
\\
\midrule
\textsc{Str-Synth} & 0.70 & 0.37 & \textbf{0.86} & \textbf{0.69}
\\
LMD (<$30$s) & 0.44 & 0.29 & \textbf{0.70} & \textbf{0.56}
\\
\midrule
LMD ($30$--$60$s) & 0.21 & 0.31 & \textbf{0.35} & \textbf{0.68}
\\
GigaMIDI & 0.57 & 0.25 & \textbf{0.61} & \textbf{0.56}
\\
NES-MDB & 0.37 & 0.28 & \textbf{0.43} & \textbf{0.61}
\\
Nottingham & 0.51 & 0.28 & \textbf{0.76} & \textbf{0.45} 

\\
\bottomrule
\end{tabular}
\label{tab:generalization}
\vspace{-0.4em}
\vspace{-0.9em}
\end{wraptable}


%% file: sections/5_related.tex
\section{Related Work}
\label{sec:related}

We discuss the most relevant literature here
and provide more discussion in Appendix~\ref{app:related}.

\paragraph{Decompilation.} 
Decompilation casts program synthesis as inferring 
source programs from non-code artifacts, including 
source code from binaries
~\citep{
tan2024llm4decompile, 
zou2025dlift, 
she2024wadec} or 
user interfaces~\citep{
wu2025ui2code, 
beltramelli2018pix2code, 
vanita2019sketch2code, 
xiao2024prototype2code}, 
vector graphics programs from raster images~\citep{
rodriguez2024starvector, 
juan2025rlrfsvg, 
belouadi2024detikzify, 
ellis2018learning}, 
and CAD programs from 3D shapes~\citep{
rukhovich2025cad, 
kolodiazhnyi2026cadrille, 
chen2025cadcrafter, 
wang2025cadgpt}. 
Earlier methods rely on SFT over paired \mbox{(artifact, code)} data~\citep{
tan2024llm4decompile, 
rodriguez2024starvector, 
rukhovich2025cad}, 
while recent work adds RL with execution-based rewards~\citep{juan2025rlrfsvg, kolodiazhnyi2026cadrille}.  
For example, \citet{juan2025rlrfsvg} 
decompile a raster image into an SVG program by 
rewarding a rendered SVG's visual similarity to the input, 
and 
\citet{kolodiazhnyi2026cadrille} 
decompile a 3D shape into a CAD program by 
rewarding 
the executed solid's geometric agreement with the input.
We extend this paradigm to music with the MIDI-to-\Strudel{} decompilation task, 
using execution-based RL that rewards faithful and readable renders.

\paragraph{Music analysis-by-synthesis.}
An observed musical signal 
can be explained by 
fitting a generative process 
whose output reconstructs it.
Prior work instantiates this idea
with parametrized 
audio-generation processes,
ranging from 
classic instrument tone analysis~\citep{risset1969analysis}
and spectral modeling synthesis~\citep{Serra1990SpectralMS}
to modern differentiable synthesis~\citep{engel2020ddsp, engel2020selfsupervised}, 
physical or instrument models~\citep{diaz2023rigid, wang2014estimation, renault2022piano}, 
and audio-effect or mixing-chain models~\citep{Steinmetz2020AutomaticMM, steinmetz2021automatic, steinmetz2022style, Steinmetz2024STITO, vanka2024diffmst, comunita2025differentiable}. 
While closely related, these methods
cast music analysis as parameter recovery for a predefined 
acoustic generator; 
\system{} instead operates over symbolic music, recovering \Strudel{} programs whose structure is inferred rather than specified a priori.

%% file: sections/6_conclusion.tex
\section{Discussion and Conclusion}
\label{conclusion}

\paragraph{Limitations and future work.}
While \system{} demonstrates that symbolic music can be decompiled into faithful and readable programs, several directions remain for future work. 
First, our readability reward uses a fixed checklist, 
which may favor abstractions that fit some musical materials better than others,
such as harmonic constructs in primarily percussive music.
Future work could make the checklist more fine-grained and input-conditional~\citep{gandhi26ppteval}. 
Second, our experiments focus on short MIDI segments; scaling to long-form music would require longer music--code data and RL reward design that captures sections, recurrence, and development. 
Third, our reward design does not explicitly optimize diversity; future work could incorporate diversity directly into reward design~\citep{tianjian2025jointly} or optimization~\citep{he2025rewarding}. Finally, music decompilation could move beyond recovering a single final program toward inferring a \emph{creative trace}: 
the evolution of a program over time that reveals how musical structure emerges through iterative REPL interactions~\citep{manesh2024sharp, hammad2026creativity}. 

In this paper, we have introduced \emph{symbolic music decompilation}, 
the task of recovering executable and editable music programs from symbolic music. 
This task requires a model to infer the higher-level musical structure that MIDI leaves implicit, 
such as repetition, layering, and temporal organization. 
We instantiated this problem as MIDI-to-\Strudel{} decompilation and presented \system{}, a two-stage post-training framework that combines supervised learning on synthetic \mbox{(\Strudel{}, MIDI)} pairs with reinforcement learning on unpaired MIDI. 
Across synthetic and real-world benchmarks, \system{} improves the faithfulness--readability tradeoff over both heuristic conversion and frontier LLMs, 
producing programs that better preserve the input music while remaining substantially more readable than note-by-note transcriptions. 
We hope this work helps facilitate future research on music decompilation systems that recover not only performed notes, but also the editable programmatic structure that makes them interpretable, reusable, and extensible.

%% file: sections/appendix.tex
\appendix

\section{\Strudel{} Background and Runtime}
\label{app:strudel}

\subsection{Program Representations}
\label{app:strudel:examples}

A central property of the 
MIDI-to-\Strudel{} task 
is that many distinct \Strudel{} programs 
can render to the same (or nearly the same) musical content (MIDI), 
while exposing different degrees of musical structure.
Figure~\ref{fig:apd:strudel_examples} 
shows four \Strudel{} programs 
that produce the same short loop of drums, bass, and electric guitar, but differ in how they represent the underlying musical ideas: 

\begin{itemize}[leftmargin=*,itemsep=1mm]
\item 
\textbf{Event-level transliteration. }
A program can stay close to the MIDI event representation by 
converting note events into fixed-resolution mini-notation patterns (Figure~\ref{fig:apd:strudel:literal}). 
At each grid step, the code emits the corresponding drum hit, pitch, or rest. 
This representation can preserve precise timing and pitches, but it produces long event lists that obscure repeated musical structure. 

\item 
\textbf{Pattern-level abstraction.}
The same material can be written more compactly 
by using pattern operations and symbolic musical units (Figure~\ref{fig:apd:strudel:mini}). 
Repeated events can be compressed into shorter patterns, 
while explicit pitch collections can be replaced with chord symbols ({\prettyfont chord()})
shaped by rhythmic operators such as {\prettyfont struct()}. This exposes rhythm and harmony hidden in a literal event list.

\item 
\textbf{Shared structure and layering.}
A program can further factor out 
musical information that is shared across voices (Figure~\ref{fig:apd:strudel:stack}). 
For example, a chord progression can be defined once and reused by different layers, so that each layer is 
expressed relative to the current chord rather than as a separate list of absolute pitches. 
Simultaneous voices can also be grouped with 
{\prettyfont stack()}, making the layered structure of the music explicit and reusable in the program.

\item 
\textbf{Arrangement-level organization.}
At a higher level, a program can separate reusable voice definitions from the temporal organization of the piece (Figure~\ref{fig:apd:strudel:arrange}). Individual parts can be named once and then combined over time using 
{\prettyfont arrange()}, allowing users to edit each layer independently and rearrange sections without rewriting the note content.

\end{itemize}

\subsection{Runtime Implementation}
\label{app:strudel:runtime}

The runtime $\mathcal{C}$ is a 
headless execution server built around the
open-source \Strudel{} implementation.\footnote{\prettyurl{https://codeberg.org/uzu/strudel}}
It provides 
two interfaces used throughout our pipeline.
First, it executes generated \Strudel{} programs and returns symbolic note
events, which are serialized to MIDI for dataset construction, reward
computation, and evaluation.
Second, it parses programs into abstract syntax trees (ASTs), which we use for
data cleaning and code-level analyses.

In implementation, 
a Node.js process loads the 
\Strudel{} JavaScript engine and 
handles program execution and AST parsing, 
while a FastAPI wrapper exposes these functions 
to Python workers and performs MIDI serialization. 
Programs that fail to compile or exceed the timeout (30s) are treated as invalid and receive the 
compilation penalty in Equation~\ref{eq:reward} during RL. 

\paragraph{MIDI construction.}
Unlike MIDI, 
a \Strudel{} program does not 
explicitly specify a finite duration.
It instead 
represents a time-queryable musical pattern~\citep{strudelpaper}:
given a requested time span, 
the runtime returns the events active 
within that span, 
including their onset, pitch, and sound name.
To obtain a finite MIDI rendering, 
$\mathcal{C}$ chooses a segment length
from the program's musical content rather than from an arbitrary query window: 
it first queries 
the pattern over a sufficiently long fixed window ($90$\,s in our experiments) 
and 
trims the resulting event sequence to one complete loop.
Specifically, 
we use the 
Knuth--Morris--Pratt algorithm~\citep{kmp_algorithm}
to identify the fundamental period 
of the queried event sequence, 
and retain only the events
within the first period.
The trimmed events are then serialized to a MIDI file by the FastAPI wrapper using the {\prettyfont mido} package~\citep{mido}.
Each \Strudel{} sound is mapped to a General MIDI program for melodic instruments or to a channel-$9$ drum note for percussion samples using fixed lookup tables~\citep{general_midi_lookup}.
Note names and numeric pitches are resolved to MIDI note numbers, 
and
\Strudel{} {\prettyfont gain} and {\prettyfont velocity} attributes 
are soft-clipped to
the MIDI velocity range.

\paragraph{AST parsing.}
For code-level analyses, 
$\mathcal{C}$ exposes a parsing interface that
returns the AST of a \Strudel{} program.
Since \Strudel{} is implemented as a JavaScript-based DSL, 
we use the Acorn
JavaScript parser~\citep{acorn} to 
parse generated programs 
into an AST before execution.
We use the resulting ASTs in two parts. 
During dataset construction (Section~\ref{sec:method:dataset}; Appendix~\ref{app:dataset}), 
the parser identifies top-level musical voices and 
traces the variable declarations on which each voice depends. 
We use this information to 
clean LLM-generated programs 
by retaining the code needed for each extracted voice 
while removing invalid fragments.
During evaluation, 
we use ASTs to compute the TSED diversity metric 
(Section~\ref{sec:exp:setup}; Appendix~\ref{app:eval:metrics}).
Specifically,
each program is serialized as a labeled tree of its syntax-node types, 
and diversity is measured by the normalized tree-edit 
distance between pairs of such trees.

\input{figures/fig_apd_strudel_examples}

\section{MIDI Representation}
\label{app:midi_repr}
This section details 
the \texttt{MidiToText} helper of Equation~\ref{eq:policy}, 
which 
converts a MIDI segment into 
the textual representation 
used as input to the language model. 
An example is shown below:

\begin{tcolorbox}[
  colback=ylightpurple,
  colframe=ylightpurple,
  width=0.99\linewidth,
  boxrule=0pt,
  arc=2pt,
  left=8pt,
  right=8pt,
  top=10pt,
  bottom=10pt,
  fontupper=\prettyfont\color{ydarkgrayazure}\linespread{1.08}\selectfont,
]
[BPM=120 Meter=4]

[acoustic\_grand\_piano] C4@0.00 E4@0.25 G4@0.50 C5@1.00

[acoustic\_bass] C2@0.00 G2@0.50 C2@1.00

[drum] bass\_drum@0.00 closed\_hi\_hat@0.25 closed\_hi\_hat@0.50 snare@0.75
\end{tcolorbox}

\paragraph{Event serialization.}
We serialize MIDI as symbolic, instrument-wise note events, 
making the input the input closer to musical notation 
and to the layered structure of \Strudel{} programs.
Specifically, 
we first parse the input MIDI 
into a flat list of note-on events, 
where each event is represented by its onset, instrument, and pitch. 
Events are then grouped by instrument
and 
serialized as a 
sequence of {\prettyfont pitch@onset} tokens sorted by onset.
For melodic instruments, 
pitches are rendered as note names (\eg {\prettyfont C4}, {\prettyfont F\#3});
for drums, 
pitches are rendered as drum names (\eg {\prettyfont closed\_hi\_hat}).
This representation choice
yields tokens closer to common textual music notation 
and idiomatic \Strudel{} code than raw MIDI indices,
while also keeping
each musical part contiguous and exposing the
segment's layered structure without requiring additional notation.

\paragraph{Cycle-based onsets.}
\Strudel{} represents
time in its native unit called \emph{cycles}
rather than seconds: 
here, 
tempo specifies 
how many cycles occur per second or minute,
and events are placed at positions within these cycles.\footnote{\prettyurl{https://strudel.cc/understand/cycles}} 
For our MIDI representation, we adopt
the convention that one cycle corresponds to 
one measure of the input segment. 
Under this convention, 
onset positions become
normalized coordinates within the measure: 
for example, in $4/4$, notes on the four beats occur at {\prettyfont 0.00}, {\prettyfont 0.25}, {\prettyfont 0.50}, and {\prettyfont 0.75}, 
instead of at absolute timestamps in seconds. 

To convert MIDI timestamps to cycle positions, we use the tempo and meter of
the segment, where BPM denotes beats per minute and meter denotes the number of
beats per measure.
Under our one-cycle-per-measure convention, one cycle spans $T_{\text{cycle}} = (60 / \text{BPM}) \times \text{meter}$
seconds. 
We then convert each MIDI onset time $t_{\text{sec}}$ to its cycle coordinate
onset $t_{\text{cycle}}$ by
\begin{equation}
    \label{eq:cycle_conversion}
    t_\text{onset}
    =
     \frac{t_{\text{sec}}}{T_{\text{cycle}}}
     =
    \frac{t_{\text{sec}}}{(60 / \text{BPM}) \times \text{meter}} .
\end{equation}

In the text representation, 
we prepend a tempo--meter header
(\eg {\prettyfont [BPM=120 Meter=4]}) and express each onset using this cycle
coordinate.
When BPM and meter metadata are available in the MIDI file, we use them directly. 
Otherwise, we render the MIDI to audio with FluidSynth~\citep{fluidsynth} and 
use \texttt{madmom}~\citep{madmom} to estimate
beat and downbeat locations, 
from which we derive BPM and meter.

\paragraph{Decompilation prompt.}
The serialized MIDI representation 
is inserted into a chat prompt for MIDI-to-\Strudel{} decompilation.
The system instruction and user message is shown below.

\begin{tcolorbox}[
  breakable, enhanced,
  title={\textbf{\system{} System Instruction}},
  colback=ylightpurple,
  colframe=ydarkblue, 
  width=0.99\linewidth,
  boxrule=1pt,
  arc=2pt,
  left=8pt,
  right=8pt,
  top=10pt,
  bottom=10pt,
  fontupper=\fontsize{9pt}{10pt}\selectfont\promptfont\color{ydarkgrayazure},
]
You are an expert live coder in Strudel, a JavaScript-based platform that ports the TidalCycles pattern language to the web. Your task is to decompile MIDI events into idiomatic Strudel code.

\promptsection{STRUDEL QUICK REFERENCE}

\promptsubsection{Basic syntax}
\begin{itemize}[leftmargin=1.2em, itemsep=0.04em, topsep=0.08em]
    \item Chain functions with dot notation, e.g., \texttt{s("bd").gain(0.8).fast(2)}.
    \item Use double quotes for mini-notation strings.
    \item Use single quotes for ordinary JavaScript strings.
    \item Use backticks for multi-line mini-notation strings.
\end{itemize}

\promptsubsection{Core functions}
\smallskip

\begin{tabularx}{\linewidth}{@{}lX@{}}
Tempo & \texttt{setcpm(bpm/bpc)}, e.g., \texttt{setcpm(120/4)} \\
Sound & \texttt{s()}, \texttt{bank()}, \texttt{n()}, \texttt{begin()}, \texttt{end()}, \texttt{loop()} \\
Pitch & \texttt{note()}, \texttt{scale()}, \texttt{chord()}, \texttt{voicing()}, \texttt{rootNotes()} \\
Timing & \texttt{fast()}, \texttt{slow()}, \texttt{off()}, \texttt{stack()}, \texttt{rev()} \\
Dynamics & \texttt{gain()}, \texttt{velocity()}, \texttt{attack()}, \texttt{delay()}, \texttt{sustain()}, \texttt{release()}, \texttt{postgain()} \\
\end{tabularx}

\smallskip
\promptsubsection{Core mini-notation}
\smallskip

\begin{tabularx}{\linewidth}{@{}lX@{}}
\texttt{"bd*4"} & Repeat four times per cycle \\
\texttt{"<bd sd>"} & Alternate across cycles \\
\texttt{"[bd sd]"} & Sequence within one cycle \\
\texttt{"bd, hh"} & Layer simultaneously \\
\texttt{"\string~"} & Silence or rest \\
\texttt{"x(5,16)"} & Euclidean rhythm with five hits over sixteen steps \\
\end{tabularx}

\promptsection{INPUT FORMAT}

\promptsubsection{Header}
\smallskip
The first line gives the tempo and meter of the source MIDI: \texttt{[BPM=120 Meter=4].}\\
Use this information to set the tempo at the top of the program: \texttt{setcpm(120/4)}.\\
If the header is missing, use \texttt{setcpm(120/4)}.

\promptsubsection{Events}
\smallskip
Each following line contains one instrument part: \texttt{[instrument\_name] pitch@cycle\_onset pitch@cycle\_onset ...}
Here, \texttt{instrument\_name} is a General MIDI program name or
\texttt{drum}; \texttt{pitch} is either a note name, such as \texttt{C4} or
\texttt{F\#3}, or a drum name, such as \texttt{bass\_drum} or
\texttt{closed\_hi\_hat}; and \texttt{cycle\_onset} is the onset position in
cycle units.
For example, in $4/4$, \texttt{0.00}, \texttt{0.25}, \texttt{0.50}, and
\texttt{0.75} correspond to the four beats of one measure.

\promptsubsection{Example}
\begin{verbatim}
[BPM=120 Meter=4]
[acoustic_grand_piano] C4@0.00 E4@0.25 G4@0.50 C5@1.00
[drum] bass_drum@0.00 closed_hi_hat@0.25 closed_hi_hat@0.50 snare@0.75
\end{verbatim}

\promptsection{TRANSLATION GUIDELINES}

\promptsubsection{1. Structure}
\begin{itemize}[leftmargin=1.2em, itemsep=0.04em, topsep=0.08em]
    \item Group instruments into separate musical layers, usually inside \texttt{stack()}.
    \item Identify repeated bars or motifs and express them with concise mini-notation.
    \item MIDI note durations are not explicitly encoded; infer reasonable durations from onset gaps and musical context.
\end{itemize}

\promptsubsection{2. Timing}
\begin{itemize}[leftmargin=1.2em, itemsep=0.04em, topsep=0.08em]
    \item Map cycle onsets directly to rhythmic subdivisions.
    \item Quantize to the smallest common subdivision that preserves the rhythm.
    \item Use \texttt{\string~} for rests, \texttt{*N} for repeated hits, \texttt{[...]} for subdivisions, and \texttt{<...>} for alternation.
\end{itemize}

\promptsubsection{3. Pitch}
\begin{itemize}[leftmargin=1.2em, itemsep=0.04em, topsep=0.08em]
    \item For melodic instruments, use \texttt{note()} with note names or \texttt{n().scale()} when a scale is clear.
    \item For drums, use \texttt{s()} with sound banks. 
    \item For same-instrument notes that occur at the same onset, prefer \texttt{chord()} with \texttt{voicing()} when the harmony is recognizable; otherwise use explicit grouped notes.
    \item Do not merge simultaneous notes from different instruments into a single chord layer.
\end{itemize}

\promptsubsection{4. Musical structure}
\begin{itemize}[leftmargin=1.2em, itemsep=0.04em, topsep=0.08em]
    \item Prefer pattern abstraction over event-by-event transliteration.
    \item Use \texttt{stack()} for simultaneous parts and \texttt{arrange()} for section-level organization.
    \item Use \texttt{<...>} for alternating cycles and \texttt{fast()} for intra-cycle density.
    \item Separate bass, chords, melody, and drums when they play distinct musical roles.
\end{itemize}

\promptsection{STYLE GUIDELINES}

\begin{itemize}[leftmargin=1.2em, itemsep=0.04em, topsep=0.08em]
    \item Ensure readability: voices, structure, and harmonic intent should be visible at a glance.
    \item Add short comments that label each major voice, such as \texttt{// walking bass} or \texttt{// backbeat drum groove}.
    \item Extract reused chord progressions, rhythms, motifs, or sounds into named \texttt{const} variables.
    \item Break long method chains across lines, with one method per line.
    \item Prefer \texttt{chord()} and \texttt{voicing()} over long explicit note lists when the harmony is clear.
    \item Prefer scale degrees with \texttt{scale()} when the line belongs to a clear key.
    \item Avoid raw MIDI numbers such as \texttt{n(60)}.
    \item Use Strudel repetition operators such as \texttt{hh*8} or \texttt{<[bar1] [bar2]>} instead of spelling out repeated events.
    \item Separate voice blocks with blank lines.
\end{itemize}

\promptsection{CRITICAL CONSTRAINTS}

\begin{enumerate}[leftmargin=1.4em, itemsep=0.06em, topsep=0.08em]
    \item \textbf{Output format:} return only raw \Strudel{} code. Do not use Markdown code fences, explanations, or conversational text.
    \item \textbf{Sound equivalence:} the rendered program should be audibly equivalent to the input MIDI in rhythm, pitch, density, and instrumentation.
    \item \textbf{Idiomatic code:} prefer clean, readable \Strudel{} patterns over verbose transliterations.
\end{enumerate}
\end{tcolorbox}

\begin{tcolorbox}[
  breakable, enhanced,
  title={\textbf{\system{} User Message Template}},
  colback=ylightpurple,
  colframe=ydarkblue, 
  width=0.99\linewidth,
  boxrule=1pt,
  arc=2pt,
  left=8pt,
  right=8pt,
  top=10pt,
  bottom=5pt,
  fontupper=\fontsize{9pt}{10pt}\selectfont\promptfont\color{ydarkgrayazure},
]
Analyze the following note events and write corresponding Strudel code.\\
\medskip
\\
\promptplaceholder{\{midi\_representation\}}
\end{tcolorbox}

During training, we vary the natural-language request in the user message while keeping the underlying MIDI representation fixed. The request is sampled from a set of paraphrased task instructions: 

\begin{quote}
\small
\promptfont
\begin{enumerate}[leftmargin=*, itemsep=0.1em, topsep=0pt]
    \item Analyze the following MIDI events and write corresponding \Strudel{} code.
    \item Can you help me turn this MIDI into \Strudel{} live coding syntax?
    \item Convert the following MIDI events into \Strudel{} code.
    \item Encode the following MIDI events into \Strudel{} notation.
    \item Express this MIDI sequence as a \Strudel{} program.
    \item Generate \Strudel{} code from the MIDI below.
    \item Generate the \Strudel{} code that reproduces the following MIDI events.
    \item Given the following MIDI events, write the equivalent \Strudel{} pattern.
    \item Interpret this MIDI and express it as \Strudel{} live coding syntax.
    \item Reproduce the following MIDI pattern in \Strudel{}.
    \item Rewrite the following MIDI events using \Strudel{} live coding syntax.
    \item Take a look at the MIDI below and write \Strudel{} code that reproduces it.
    \item Transform this MIDI representation into \Strudel{}.
    \item Translate this MIDI sequence into \Strudel{} code.
    \item Write \Strudel{} code that is functionally equivalent to the following MIDI.
\end{enumerate}
\end{quote}

\section{Descriptions for Optimization Methods}
\label{app:grpogdpo}

We provide a description of the two policy-optimization methods
referenced in the main paper (Section~\ref{sec:method:training}). 
We first explain Group Relative Policy Optimization (GRPO) in Section~\ref{app:grpogdpo:grpo} 
and 
Group reward-Decoupled normalization Policy Optimization (GDPO) in Section~\ref{app:grpogdpo:gdpo}.
For detailed explanations and ablations, 
please refer to the original papers~\citep{guo2025deepseek, gdpo}.

We consider a general verifiable generation setting 
in which a policy $\pi_\theta$ maps 
an input $x$ to an output $\hat{y}$, 
and an external evaluator assigns rewards after generation.
The evaluator may be non-differentiable, 
since the policy-gradient objectives below 
only require scalar reward values and 
do not differentiate through the evaluation procedure.
This setting covers our training setup, where rewards are
computed from the executed output and auxiliary evaluation metrics.

\subsection{Group Relative Policy Optimization (GRPO)}
\label{app:grpogdpo:grpo}
GRPO~\citep{guo2025deepseek} is a
policy-gradient method
that estimates advantages relative to a \emph{group}
of completions sampled for the same input prompt,
removing the separately trained value network of PPO~\citep{schulman2017ppo}.
For each input $x$, 
the old policy samples $G$ candidates $\{\hat{y}^{(g)}\}_{g=1}^{G}\sim\pi_{\theta_{\text{old}}}(\cdot\mid x)$.
Each candidate is scored 
with a scalar reward $r^{(g)}=r(x,\hat{y}^{(g)})$, 
and the rewards are normalized within the group:
\begin{equation}
\label{eq:app:grpo-adv}
    A^{(g)} = \frac{r^{(g)} - \mu}{\sigma + \varepsilon},
    \quad
    \mu = \frac{1}{G}\sum_{g=1}^{G} r^{(g)},
    \quad
    \sigma = \mathrm{std}\!\left(\{r^{(g)}\}_{g=1}^{G}\right).
\end{equation}

The policy is then updated 
by maximizing 
a token-level clipped surrogate 
with a KL penalty to a reference policy $\pi_{\text{ref}}$:
\begin{align}
\label{eq:app:grpo}
    \mathcal{J}_{\text{GRPO}}(\theta)
    &= \mathbb{E}\!\left[\frac{1}{G}\sum_{g=1}^{G}
        \frac{1}{|\hat{y}^{(g)}|} \sum_{t=1}^{|\hat{y}^{(g)}|}
        \min\!\left(
            \rho_t^{(g)} A^{(g)},\;
            \mathrm{clip}\!\left(\rho_t^{(g)},\, 1{-}\epsilon,\, 1{+}\epsilon\right) A^{(g)}
        \right)\right] \nonumber \\
    &\quad - \beta\,
    \mathbb{D}_{\text{KL}}\!\left[\pi_\theta \,\big\|\, \pi_{\text{ref}}\right],
\end{align}

where $t$ indexes positions in the generated token sequence and
\begin{equation}
\rho_t^{(g)}
=
\frac{
\pi_\theta(\hat{y}_t^{(g)}\mid \hat{y}_{<t}^{(g)}, x)}
{\pi_{\theta_{\mathrm{old}}}(\hat{y}_t^{(g)}\mid \hat{y}_{<t}^{(g)}, x)}
\end{equation}
denotes the per-token importance ratio between the current and old policies.
Here, $\epsilon$ is the clipping threshold, 
$\beta$ is the KL coefficient,
and $\mathbb{D}_{\text{KL}}\!\left[\pi_\theta \,\big\|\, \pi_{\text{ref}}\right]$ 
denotes the
averaged token-level KL regularization term.

\subsection{Group reward-Decoupled normalization Policy Optimization (GDPO)}
\label{app:grpogdpo:gdpo}

GRPO assumes a scalar reward for each sampled output;
when the training signal has multiple reward components, 
a straightforward way to apply GRPO is to 
first scalarize them and then apply the 
group-wise normalization in Equation~\ref{eq:app:grpo-adv}.
Concretely, for each sampled output $\hat{y}^{(g)}$, one may form
\begin{equation}
\label{eq:app:wsum}
   r^{(g)} = \sum_{k=1}^{K} w_k r_k^{(g)}, 
\end{equation}
where $r_k^{(g)}=r_k(x,\hat{y}^{(g)})$ 
is the $k$-th reward component and $w_k$ is its corresponding weight.
Because normalization is applied only after scalarization, 
high-variance components can dominate 
both the scalar reward and the resulting advantage, even when the intended objective assigns non-negligible weight to multiple components.

GDPO~\citep{gdpo} instead 
normalizes each reward component 
within the group
before aggregation. 
For each component $k$, 
GDPO computes a component-wise group-relative advantage
\begin{equation}
\label{eq:app:gdpoadv}
A_k^{(g)}
=
\frac{r_k^{(g)}-\mu_k}{\sigma_k+\varepsilon},
\quad
\mu_k=\frac{1}{G}\sum_{g=1}^{G} r_k^{(g)},
\quad
\sigma_k = \mathrm{std}\bigl(\{r_k^{(g)}\}\bigr),
\end{equation}

aggregates the normalized advantages with the weights $w_k$:
\begin{equation}
\label{eq:app:gdpowsum}
A_{\text{sum}}^{(g)}
=
\sum_{k=1}^{K} w_k A_k^{(g)},
\end{equation}
and finally applies a batch-wise normalization across all rollouts in the
batch $\mathcal{B}$ to keep the advantage magnitude from growing with the number
of objectives:
\begin{equation}\label{eq:gdpo-batchnorm}
\hat{A}_{\text{sum}}^{(g)}
= \frac{A_{\text{sum}}^{(g)} - \mathrm{mean}\bigl\{A_{\text{sum}}^{(g')} \,\big|\, g' \in \mathcal{B}\bigr\}}
{\mathrm{std}\bigl\{A_{\text{sum}}^{(g')} \,\big|\, g' \in \mathcal{B}\bigr\} + \varepsilon}.
\end{equation}

The policy is updated with 
the clipped surrogate of Equation~\ref{eq:app:grpo}, 
with
$A^{(g)}$ replaced by $\hat{A}_{\text{sum}}^{(g)}$.

\section{Readability Reward Design}
\label{app:rubrics}

The readability reward $R_{\text{read}}$ (Section~\ref{sec:method:reward})
is computed by an LLM judge 
using a fixed binary checklist of 
$J=12$ items.
We use rule-based binary criteria
to obtain a stable reward signal in a domain where many programs
can be valid decompilations, 
following the rubric-as-rewards paradigm~\citep{gunjal2026rubrics, viswanathan2025checklists, gandhi26ppteval}.
The checklist combines general 
code readability criteria, 
adapted from prior readability metrics~\citep{bnw, scalabrino2018readability}, 
with
\Strudel{}-specific criteria for concision and musical abstraction. 
Specifically, 
items $1$--$5$ capture general readability, such as variable reuse, comments, and formatting, 
while 
items $6$--$12$ reward the use of 
\Strudel{}'s pattern operators and musical abstractions
over note-by-note transliteration. 
The
score is the fraction of satisfied items: 
$R_{\text{read}}=\tfrac{1}{12}\sum_j c_j \in [0,1]$.

We use \texttt{Qwen3.6-35B-A3B}~\citep{qwen36_35b_a3b} as the judge. 
We follow the Qwen recommended temperature setting of $0.7$, 
set the maximum output length to $4000$ tokens, 
and disable thinking ({\prettyfont enable\_thinking=false}). 
The full judge system instruction, including the rubric, is shown below.

\begin{tcolorbox}[
  breakable, enhanced,
  title={\textbf{Readability Judge System Instruction}},
  colback=ylightpurple,
  colframe=ydarkblue, 
  width=0.99\linewidth,
  boxrule=1pt,
  arc=2pt,
  left=8pt,
  right=8pt,
  top=10pt,
  bottom=10pt,
  fontupper=\fontsize{9pt}{10pt}\selectfont\promptfont\color{ydarkgrayazure},
]
You are an expert Strudel code reviewer. 
Strudel is a JavaScript-based live coding language for music. 
Given a Strudel program, your task is to evaluate its readability using the 12 binary checklist items below. Apply each criterion literally and independently. 
Return only valid JSON in the following format: 
\begin{verbatim}
{"items":[{"id":"1","reasoning":"...","score":0},...,{"id":"12",...}]}
\end{verbatim}

Include all 12 items in order. Each score must be exactly 0 or 1. Keep each reasoning to one short sentence.

\medskip
\textbf{Item 1: Variable usage.} 
Score $1$ iff the code declares 
at least one top-level \texttt{const X = \dots} 
or \texttt{let X = \dots} and reuses \texttt{X} elsewhere.

\medskip
\textbf{Item 2: Comment usage.}
A meaningful comment is a \texttt{//} or \texttt{/* */} comment with at least two words (\eg, \texttt{// syncopated drum groove}). Decoration-only comments do not count. Score $1$ iff $\text{meaningful\_comments}/\max(\text{code\_lines},1)\ge 0.10$.

\medskip
\textbf{Item 3: Repeated method chains.}
Score $0$ iff the code repeats the same contiguous sequence of $\ge 4$ chained dot-method calls at least twice. 
The repeated sequence must match exactly 
in both method names and arguments. 
Otherwise score $1$. 
Do not count expression heads such as \texttt{note(\dots)}, \texttt{stack(\dots)}, or \texttt{chord(\dots)}.

\medskip
\textbf{Item 4: Line length.}
Over non-empty lines, score $1$ iff the average line length is at least $10$ characters and the maximum line length is at most $90$ characters.

\medskip
\textbf{Item 5: Blank line usage.}
Score $1$ iff blank lines make up at least $10\%$ of all code lines, i.e., $\text{blank\_lines}/\max(\text{code\_lines},1)\ge 0.10$.

\medskip
\textbf{Item 6: Mini-notation compression.}
Inspect each mini-notation string passed to \texttt{note}, \texttt{n}, \texttt{sound}, or \texttt{s}. 
Score $0$ iff it contains any avoidable repetition: 
the same token/group appears four or more times in a row, 
a phrase repeats in a row, 
or a \texttt{<\dots>} alternation repeats the same element. 
Ignore repetitions already expressed with \texttt{*N} or \texttt{!N}. Otherwise score $1$.

\medskip
\textbf{Item 7: Time transformation.}
Score $1$ iff the code uses at least one timing-change methods: \texttt{.fast}, \texttt{.slow}, \texttt{.ply}, \texttt{.iter}, \texttt{.early}, \texttt{.late}.

\medskip
\textbf{Item 8: Rhythm structure..}
Score $1$ iff the code uses at least one of rhythm-structure method:  \texttt{.struct}, \texttt{.mask}, \texttt{.euclid}, \texttt{.euclidLegato}, \texttt{.euclidRot}, \texttt{.swing}, \texttt{.swingBy}.

\medskip
\textbf{Item 9: Pattern layering.}
Score $1$ iff the code uses at least one of pattern-layering method: \texttt{.jux}, \texttt{.layer}, \texttt{.superimpose}.

\medskip
\textbf{Item 10: Scale usage.}
Score $1$ iff the code uses chained \texttt{.scale(\dots)} to map scale degrees to pitches
(e.g., \verb+n("0 2 4").scale("c:major")+).

\medskip
\textbf{Item 11: Chord voicing.}
Score $1$ iff the code uses \texttt{chord(\dots)} with a voicing-related method in the same expression chain:
\texttt{.voicing(\dots)} or \texttt{.rootNotes(\dots)}.

\medskip
\textbf{Item 12: Pitch shift.}
Score $1$ iff the code uses at least one pitch-shift operation:
\texttt{.transpose(\dots)}, \texttt{.add(note(\dots))}, \texttt{.sub(note(\dots))}, or \texttt{.add}/\texttt{.sub} with an integer semitone count. Small-float \texttt{.add(<1)} for detuning and non-pitch strings do not count.

\end{tcolorbox}

\section{\dataset{}}
\label{app:dataset}

This section provides additional details on the construction of \dataset{}, 
the synthetic paired corpus used to train \system{} (Section~\ref{sec:method:dataset}). 
We describe the overall generation pipeline in
Section~\ref{app:dataset:pipeline} 
and 
the conditioning inputs 
used to
control musical content and coding style of the generated \Strudel{} programs in
Section~\ref{app:dataset:conditioning}.

\subsection{Generation Pipeline Details}
\label{app:dataset:pipeline}

We detail each stage of the data generation pipeline used to construct \dataset{}.

\paragraph{Stage 1: Conditioned generation.}
We prompt \claude{}~\citep{anthropic2026opus46} with a system prompt 
instructing it to write a complete, idiomatic \Strudel{} program, conditioned on
a musical seed $s_{\text{music}}$ and a code-style seed $s_{\text{code}}$ supplied
in the user message 
(detailed in Appendix~\ref{app:dataset:conditioning}).
To ensure deterministic MIDI rendering, 
we forbid 
random modifiers 
and constructs unsupported by the headless engine, including visuals, sliders, and external samples.
We sample with temperature $1.0$ and up to $4{,}096$ new tokens, 
with extended thinking disabled. 
The system prompt is shown below.

\begin{tcolorbox}[
  breakable, enhanced,
  title={\textbf{\dataset{} Generation System Instruction}},
  colback=ylightpurple,
  colframe=ydarkblue, 
  width=0.99\linewidth,
  boxrule=1pt,
  arc=2pt,
  left=8pt,
  right=8pt,
  top=10pt,
  bottom=10pt,
  fontupper=\fontsize{9pt}{10pt}\selectfont\promptfont\color{ydarkgrayazure},
]
You are an expert live coder specializing in Strudel, a JavaScript-based
platform that ports the TidalCycles pattern language to the web. Your
task is to translate natural language music descriptions into executable, idiomatic Strudel code.

\promptsection{STRUDEL QUICK REFERENCE}

\promptsubsection{Basic syntax}
\begin{itemize}[leftmargin=1.2em, itemsep=0.05em, topsep=0.1em]
    \item Chain functions with dot notation.
    \item Use double quotes (\texttt{"}) for single-line mini-notation patterns.
    \item Use backticks (\texttt{`}) for multi-line mini-notation strings.
    \item Use single quotes (\texttt{'}) for regular strings that are not parsed as mini-notation.
\end{itemize}

\promptsubsection{Core mini-notation} 
\smallskip

\begin{tabularx}{\linewidth}{@{}lX@{}} 
\texttt{"bd*4"} & Repeat four times per cycle \\ 
\texttt{"<bd sd>"} & Alternate across cycles \\ 
\texttt{"[bd sd]"} & Sequence within one cycle \\ 
\texttt{"bd, hh"} & Layer simultaneously \\ 
\texttt{"\string~"} & Silence or rest \\ 
\texttt{"bd(3,8)"} & Euclidean rhythm shorthand 
\\ \texttt{"c@3 e"} & elongate a note with \texttt{@} \\ 
\end{tabularx}

\smallskip
\promptsubsection{Core functions} 
\smallskip

\begin{tabularx}{\linewidth}{@{}lX@{}} 
Tempo & \texttt{setcpm(bpm/bpc)}, e.g., \texttt{setcpm(120/4)} \\ 
Sound & \texttt{s()}, \texttt{bank()}, \texttt{n()}, \texttt{begin()}, \texttt{end()}, \texttt{loop()} \\ 
Pitch & \texttt{note()}, \texttt{scale()}, \texttt{chord()}, \texttt{voicing()}, \texttt{rootNotes()} \\ 
Timing & \texttt{fast()}, \texttt{slow()}, \texttt{off()}, \texttt{stack()}, \texttt{arrange()}, \texttt{rev()}, \texttt{jux()} \\ 
Rhythm & \texttt{struct()}, \texttt{mask()}, \texttt{ply()}, \texttt{euclid()}, \texttt{swingBy()} \\ 
Dynamics & \texttt{gain()}, \texttt{velocity()}, \texttt{attack()}, \texttt{sustain()}, \texttt{release()}, \texttt{adsr()}, \texttt{postgain()} \\ 
Modulation & \texttt{fm()}, \texttt{fmh()}, \texttt{vib()} \\ 
Filters & \texttt{lpf()}, \texttt{hpf()}, \texttt{bpf()}, \texttt{bpq()} \\ 
Effects & \texttt{room()}, \texttt{delay()}, \texttt{chorus()}, \texttt{compressor()} \\ 
\end{tabularx} 

\smallskip
\promptsubsection{General advice} 
\begin{itemize}[leftmargin=1.2em, itemsep=0.04em, topsep=0.08em] 
\item Use \texttt{\$:} labeled statements or \texttt{stack()} to layer simultaneous parts. 
\item Use \texttt{arrange()} for song-level section structure. 
\item Use \texttt{voicing()} for chord progressions and \texttt{rootNotes(n)} for bass lines from chord symbols. 
\item Use \texttt{struct()} for rhythmic gating; use \texttt{off()} or \texttt{superimpose()} for harmonic echoes. 
\item Use Euclidean rhythms \texttt{(n,k)} and \texttt{swingBy()} for natural-feeling grooves. 
\item Use synths such as \texttt{sawtooth}, \texttt{square}, \texttt{sine}, and \texttt{triangle}, or GM soundfonts such as \texttt{gm\_acoustic\_bass} and \texttt{gm\_electric\_piano\_1}. 
\end{itemize} 

\promptsection{INPUT FORMAT: MUSIC DESCRIPTION} 

The input is a natural-language music description specifying any combination of genre, mood, instrumentation, tempo, key, time signature, or structural ideas. Descriptions range from terse prompts such as ``chill lo-fi beat'' to detailed requests. Treat missing details as creative latitude. 

\promptsection{GENERATION GUIDELINES} 

\promptsubsection{1. Interpret the description} 

\begin{itemize}[leftmargin=1.2em, itemsep=0.04em, topsep=0.08em] 
\item Infer genre conventions: tempo, instruments, rhythmic feel, and harmonic language.  
\item Plan the main sound layers from the description, such as melody, drums, bass, and pads.
\end{itemize} 

\promptsubsection{2. Build harmony} 

\begin{itemize}[leftmargin=1.2em, itemsep=0.04em, topsep=0.08em] 
\item Express chord progressions symbolically, e.g., \texttt{chord("<Am7 Fmaj7 G C>").voicing()}. 
\item Control register with anchor (e.g., \texttt{.anchor("c4")}) and mode \\(e.g., \texttt{.mode("below")}, \texttt{.mode("duck")}, or \texttt{.mode("above")}). 
\item Derive basslines using chord symbols, e.g., \\ \texttt{chord(chords).rootNotes(2).s("gm\_acoustic\_bass")}. 
\item Use interval stacks such as \texttt{note("<[c3,e3,g3] [a2,c3,e3]>")} for explicit voicings. 
\end{itemize} 

\promptsubsection{3. Build rhythm and structure} 

\begin{itemize}[leftmargin=1.2em, itemsep=0.04em, topsep=0.08em] 
\item Use \texttt{stack()} or \texttt{\$:} for simultaneous parts, and \texttt{arrange([n, section], ...)} \\for song sections. 
\item Use Euclidean rhythms for organic feels, e.g., \texttt{s("bd(3,8)")} or \texttt{s("hh(5,8)")}. 
\item Add swing with \texttt{.swingBy(1/8, 4)} or \texttt{.swing(4)}. 
\item Add variation with pattern effects, e.g., \texttt{.add("<0 1 2 1>")}, or \texttt{.off(1/8, add(7))}. 
\end{itemize} 

\promptsubsection{4. Apply style and effects} 

\begin{itemize}[leftmargin=1.2em, itemsep=0.04em, topsep=0.08em] 
\item Match effects to genre, such as reverb for ambient or jazz, compression for electronic music, and delay for dub. 
\item Keep patterns concise by using mini-notation repetition and \texttt{<...>} alternation instead of spelling out every bar. 
\item Use \texttt{layer()} or \texttt{superimpose()} to add harmonic richness without writing extra voices manually. 
\end{itemize} 

\promptsection{FORBIDDEN CONSTRUCTS}

The following constructs must not be used, even if they appear in a few-shot example.

\begin{itemize}[leftmargin=1.2em, itemsep=0.08em, topsep=0.08em]

\item \textbf{Visuals:}\\
{\prettyfont .color()},
{\prettyfont .\_scope()},
{\prettyfont .\_pianoroll()},
{\prettyfont .\_waveform()},
{\prettyfont Hydra},
{\prettyfont initHydra},
{\prettyfont src()}, {\prettyfont .out()}

\item \textbf{Randomness:}\\
{\prettyfont rand},
{\prettyfont irand},
{\prettyfont perlin},
{\prettyfont Math.random()},
{\prettyfont choose},
{\prettyfont wchoose},
{\prettyfont chooseCycles},
{\prettyfont randcat},\\
{\prettyfont wchooseCycles},
{\prettyfont wrandcat}

\item \textbf{Probabilistic modifiers:}\\
{\prettyfont degradeBy},
{\prettyfont degrade},
{\prettyfont undegradeBy},
{\prettyfont undegrade},
{\prettyfont sometimesBy},
{\prettyfont sometimes},
{\prettyfont someCyclesBy},
{\prettyfont someCycles},
{\prettyfont often},
{\prettyfont rarely},
{\prettyfont almostNever},
{\prettyfont almostAlways},
{\prettyfont never},
{\prettyfont always}, 
{\prettyfont "bd?"}, 
{\prettyfont "bd | sd"}

\item \textbf{Interactivity:}\\
{\prettyfont slider()},
{\prettyfont button()},
{\prettyfont xy()},
{\prettyfont midin()},
{\prettyfont cc()},
{\prettyfont initMidi()},
{\prettyfont await midin()}

\item \textbf{External samples:}\\
{\prettyfont loadSamples()},
{\prettyfont samples('url')},
and any explicit {\prettyfont .wav} or {\prettyfont .mp3} path string

\end{itemize}

\promptsection{CRITICAL CONSTRAINTS} 

\begin{enumerate}[leftmargin=1.4em, itemsep=0.06em, topsep=0.08em] 
\item \textbf{Output format:} return only raw \Strudel{} code. Do not use Markdown code fences, conversational text, or explanations. The output must be directly copy-pasteable into the \Strudel{} REPL. 
\item \textbf{Fidelity:} closely follow the musical intent of the description. 
\item \textbf{Idiomatic code:} prefer clean, readable \Strudel{} patterns with mini-notation abstractions rather than verbose literal sequences. 
\item \textbf{Tempo:} always use \texttt{setcpm(bpm/bpc)}, e.g., \texttt{setcpm(120/4)} or \texttt{setcpm(90/2)}. 
\end{enumerate} \end{tcolorbox}

\paragraph{Stage 2: Cleaning and dead-voice pruning.}
LLM-written programs frequently contain \emph{dead voices}: music layers that compile successfully but emit no audible events. 
These typically arise from hallucinated instrument names or malformed operator chains.
For example, 
if a voice in the generated program 
references 
a sound name that does not exist in the \Strudel{} sound library, 
such as {\prettyfont \$: note("c\# e g a").sound("sound\_404")}, 
the program may still compile during rendering (Stage~3), 
but that voice would not produce any audible events. 
Such dead voices are problematic for training:
the code suggests an active musical layer, while the rendered MIDI contains nothing, creating a mismatch between the program and the music that the model would otherwise learn as a valid pattern.
We address this in two passes: regex rewriting and voice-level pruning.

We first apply 
a fixed sequence of deterministic regular expression rewrites
that correct recurring syntactic mistakes we observed in early generations.
The rewrites used in our pipeline are summarized in Table~\ref{tab:apd:regex_fixes};
together, they correct ${\sim}37\%$ of programs at least once before any compilation is attempted\footnote{These rewrites are heuristic and target the most common failure modes observed. They are not guaranteed to be sound and may occasionally modify already-valid programs; cleaned outputs are re-validated afterward.}.

\input{tables/regex_table}

We then parse each rewritten program 
with Acorn JavaScript parser~\citep{acorn} 
and use the resulting AST to identify top-level musical voices, 
following the representative \Strudel{} structures 
introduced earlier in Appendix~\ref{app:strudel:examples}: 
mini-notation statements 
({\prettyfont \$:}), {\prettyfont stack} arguments, 
and {\prettyfont arrange} tracks 
(Figures~\ref{fig:apd:strudel:mini}-\ref{fig:apd:strudel:arrange}).
For each voice, 
we record its character span in the original source 
and collect the top-level variable declarations
(\eg {\prettyfont let chords = chord(``<Bbm9 Fm9>/4'')})
and helper definitions 
(\eg {\prettyfont samples(`github:eddyflux/crate')})
it depends on. 
We render this isolated program 
with the runtime $\mathcal{C}$ using a $60$\,s window, 
and remove the voice if it produces zero audible events.
Finally, we remove top-level declarations 
that are no longer referenced and retain a program only if at least one voice survives. 
This full parsing-and-pruning procedure is illustrated in 
Figure~\ref{fig:apd:data_cleaning}.

\input{figures/fig_apd_data_cleaning}

\paragraph{Stage 3: Rendering.}
Each cleaned program is rendered to MIDI 
with the runtime $\mathcal{C}$, 
which queries the pattern over a fixed window ($90$\,s) 
and trims the result to one fundamental loop (Appendix~\ref{app:strudel:runtime}).
We discard 
renderings shorter than $2$\,s, 
which are typically degenerate programs 
that produce a very short event sequence.
The surviving \mbox{(MIDI, \Strudel{})} pairs constitute \dataset{}; summary statistics are reported in Table~\ref{tab:dataset_stats}.

\subsection{Conditioning Inputs}
\label{app:dataset:conditioning}

We condition each generation on two inputs: a musical seed $s_{\text{music}}$ that specifies what kind of piece to write, and a code-style seed $s_{\text{code}}$ that specifies how the piece should be expressed as a \Strudel{} program. 
These two seeds vary independently 
so that the same musical target 
can be realized in different program idioms 
and a given idiom can be applied to diverse music. 

\input{tables/music_seed_table}

\paragraph{Musical seed $s_{\text{music}}$.}
The musical seed specifies 
the intended musical content of 
each generated piece 
and is injected into the user message of the generation prompt.
We construct these seeds from three metadata sources: 
(i) EveryNoise~\citep{mcdonald_everynoise1d}, a fine-grained vocabulary of genres from Spotify listening data; 
(ii) TheoryTab~\citep{theorytab}, a collection of popular song analyses;
and
(iii) MidiCaps~\citep{midicaps}, a captioned MIDI dataset built on LMD~\citep{lakhmidi} that combines MIR-extracted metadata with LLM-generated natural-language descriptions.
To control how much musical information 
is given to the LLM generator, 
we use prompts at several levels of musical detail, from genre-only prompts to richer prompts that include multiple musical attributes. Table~\ref{tab:apd:music_seed} summarizes these conditioning modes and their exposed attributes.

\paragraph{Code-style seed $s_{\text{code}}$.}
The code-style seed selects 
among 
program idioms 
that realize similar musical targets 
with different code structure: 
(i) labeled mini-notation patterns using
{\prettyfont \$:}; 
(ii) layered programs built with {\prettyfont stack};
and 
(iii) multi-section arrangements using {\prettyfont arrange}
(see Figures~\ref{fig:apd:strudel:mini}--\ref{fig:apd:strudel:arrange} for examples). 
To further ground \dataset{} in real \Strudel{} practice, 
we additionally draw
structural templates from 
web-crawled programs 
and provide them as references for a randomly selected subset of generations.
Specifically, using the AST-based procedure in Appendix~\ref{app:dataset:pipeline}, 
we parse
each program into top-level musical voices, 
sample one or two voices, 
and include
the resulting fragments as a reference for
\Strudel{} coding patterns.
These templates are used only to 
guide code structure and operator use; 
the model is still instructed to compose new musical content.
Each code-style seed is implemented with a style-specific user prompt with an optional reference-template block for template-conditioned generations; we list the prompts below.

\begin{tcolorbox}[
  breakable, enhanced,
  title={\textbf{\dataset{} Generation User Prompt: (1) \texttt{\$:} label syntax}},
  colback=ylightpurple,
  colframe=ydarkblue, 
  width=0.99\linewidth,
  boxrule=1pt,
  arc=2pt,
  left=8pt,
  right=8pt,
  top=5pt,
  bottom=10pt,
  fontupper=\fontsize{9pt}{10pt}\selectfont\promptfont\color{ydarkgrayazure},
]
\promptsection{STRUCTURE}
Your output's top-level structure must use labeled \Strudel{} statements of the form \texttt{\$:} \texttt{pattern}, with one statement per musical layer.

\smallskip
\begin{itemize}[leftmargin=1.2em, itemsep=0.04em, topsep=0.08em] 
\item Use $3$--$5$ top-level \texttt{\$:} statements, corresponding to musical layers such as drums, bass, chords, melody, or pads. 
\item Each layer should be written primarily with concise mini-notation patterns. 
\item Do not use a top-level \texttt{stack} or \texttt{arrange} call; keep layers as separate \texttt{\$:} statements. 
\item Use idiomatic pattern operations such as \texttt{struct()}, \texttt{fast()}, \texttt{slow()}, \texttt{off()}, \texttt{ply()}, \texttt{swingBy()}, \texttt{chord(...).voicing()}, and \texttt{n(...).scale(...)} when useful. 
\item Keep the program readable by reusing variables for shared chord progressions, scales, rhythms, or sound choices.
\end{itemize}

\medskip
\emph{\{\% Optional block: included only when a reference program is provided.\%\}}

\promptsection{REFERENCE TEMPLATE} 
Use the following reference only as a guide for coding style and program organization. 
Write entirely new musical content from scratch, including your own choice of key, harmony, rhythm, and instrumentation.

\promptplaceholder{\{reference\_code\}}

The reference may contain forbidden constructs, such as randomness, visuals, interactivity, or external samples. Do not reproduce those constructs; use deterministic and headless-compatible alternatives instead.

\smallskip
\emph{\{\% End optional block. \%\}}

\medskip
Create music in \Strudel{} live coding language using the following specifications:

\promptplaceholder{\{music\_attributes\}}
 
\end{tcolorbox}

\begin{tcolorbox}[
  breakable, enhanced,
  title={\textbf{\dataset{} Generation User Prompt: (2) \texttt{stack}}},
  colback=ylightpurple,
  colframe=ydarkblue, 
  width=0.99\linewidth,
  boxrule=1pt,
  arc=2pt,
  left=8pt,
  right=8pt,
  top=5pt,
  bottom=10pt,
  fontupper=\fontsize{9pt}{10pt}\selectfont\promptfont\color{ydarkgrayazure},
]
\promptsection{STRUCTURE}
Your output's top-level structure must be a single \texttt{stack(...)} call containing $3$--$5$ inner streams.

\smallskip
\begin{itemize}[leftmargin=1.2em, itemsep=0.04em, topsep=0.08em] 
\item Each stream should loop within $4$--$8$ cycles. 
\item Inside each stream, you may use any idiomatic \Strudel{} pattern freely. 
\item Do not use top-level \texttt{\$:} statements; wrap all streams inside a single \texttt{stack(...)} instead. 
\item Use clear variable names and add comments when they improve readability. 
\end{itemize}

\medskip
\emph{\{\% Optional block: included only when a reference program is provided.\%\}}

\promptsection{REFERENCE TEMPLATE} 
Use the following reference only as a guide for coding style and program organization. 
Write entirely new musical content from scratch, including your own choice of key, harmony, rhythm, and instrumentation.

\promptplaceholder{\{reference\_code\}}

The reference may contain forbidden constructs, such as randomness, visuals, interactivity, or external samples. Do not reproduce those constructs; use deterministic and headless-compatible alternatives instead.

\smallskip
\emph{\{\% End optional block. \%\}}

\medskip
Create music in \Strudel{} live coding language using the following specifications:

\promptplaceholder{\{music\_attributes\}} 
 
\end{tcolorbox}

\begin{tcolorbox}[
  breakable, enhanced,
  title={\textbf{\dataset{} Generation User Prompt: (3) \texttt{arrange}}},
  colback=ylightpurple,
  colframe=ydarkblue, 
  width=0.99\linewidth,
  boxrule=1pt,
  arc=2pt,
  left=8pt,
  right=8pt,
  top=5pt,
  bottom=10pt,
  fontupper=\fontsize{9pt}{10pt}\selectfont\promptfont\color{ydarkgrayazure},
]
\promptsection{STRUCTURE}
Your output's top-level structure must be a \texttt{arrange([cycles, section\_expr], ...)} call.

\smallskip
\begin{itemize}[leftmargin=1.2em, itemsep=0.04em, topsep=0.08em] 
\item Use $2$--$4$ sections. 
\item Each section should last $2$--$3$ cycles, with a total length of at most $12$ cycles. 
\item Each \texttt{section\_expr} could be any idiomatic Strudel expression, including \texttt{stack(...)}, mini-notation, scale-degree melodies, symbolic chord progressions, drum banks, polyrhythms, and effects.
\item Use clear variable names and add comments when they improve readability. 
\end{itemize}

\medskip
\emph{\{\% Optional block: included only when a reference program is provided.\%\}}

\promptsection{REFERENCE TEMPLATE} 
Use the following reference only as a guide for coding style and program organization. 
Write entirely new musical content from scratch, including your own choice of key, harmony, rhythm, and instrumentation.

\promptplaceholder{\{reference\_code\}}

The reference may contain forbidden constructs, such as randomness, visuals, interactivity, or external samples. Do not reproduce those constructs; use deterministic and headless-compatible alternatives instead.

\smallskip
\emph{\{\% End optional block. \%\}}

\medskip
Create music in \Strudel{} live coding language using the following specifications:

\promptplaceholder{\{music\_attributes\}}
 
\end{tcolorbox}

\section{Evaluation Details}
\label{app:eval}

\subsection{Metric Definitions}
\label{app:eval:metrics}

We organize evaluation metrics 
along the three axes of faithfulness, readability, and diversity introduced in Section~\ref{sec:exp:setup}.
The main result table (Table~\ref{tab:main}) reports the metric mean over the evaluation set under a single generation per input.
For the generalization datasets (Section~\ref{sec:exp:generalization}, Table~\ref{tab:generalization}) 
and additional evaluations (Appendix~\ref{app:more_eval}), 
we instead report Best@$K$ ($K=5$): 
each method generates five candidate programs per input, and we select the candidate with the highest Inst F1. All remaining metrics are computed on the candidates selected by Inst F1.

\paragraph{Faithfulness.}
We measure how accurately the rendered MIDI $\hat{x}=\mathcal{C}(\hat{y})$ reproduces the input MIDI $x$,
adapting metrics from the automatic music transcription literature~\citep{mt3, maman2022unaligned, yourmt3, shin2026music}
together with executability check 
from decompilation literature~\citep{tan2024llm4decompile, wong2023refining, armengol2024slade}.

\begin{itemize}[leftmargin=*,itemsep=0.3mm]
\item \textbf{Comp.\ (compile rate)} 
is the fraction of generated programs that the runtime $\mathcal{C}$ successfully executes to a non-empty event list within the $30$\,s timeout.
Programs that error out, time out, or emit zero events are counted as failures.

\item \textbf{Onset F1} treats 
each note as a \mbox{(pitch, onset)} tuple
and counts a predicted note as correct 
if it has the same pitch 
as a reference note 
whose onset lies within $\pm 50$\,ms of the prediction.
We use the standard \texttt{mir\_eval} implementation~\citep{mireval} to perform optimal bipartite matching between reference and predicted notes, 
and compute per-piece precision, recall, and F1 from the resulting match counts. 
We report Onset F1 as the primary metric, following evidence that onset correctness is the principal determinant of perceptual transcription accuracy~\citep{perceptualonset}.

\item \textbf{Frame F1} measures pianoroll-level overlap of sustained notes.
A pianoroll is a $[T \times 128]$ binary matrix
where the entry at frame $t$ and pitch $p$ is $1$ 
iff at least one note of pitch $p$ is sounding during the $10$\,ms span covered by frame $t$
(\ie we use a frame rate of $100$ frames per second); 
each note's velocity is binarized so that any non-zero velocity counts as active. 
Frame F1 is the element-wise 
binary F1 between the reference 
and predicted pianorolls,
computed using \texttt{precision\_recall\_fscore\_support}
from \texttt{scikit-learn}~\citep{scikit-learn}.

\item \textbf{Inst F1 (multi-instrument onset F1)} 
is the same as Onset F1 but 
with notes partitioned 
by their assigned General MIDI program number,
treating each program as a distinct instrument
and the channel-$9$ percussion track as a separate drum partition.
Each partition is scored independently with the $\pm 50$\,ms onset criterion,
and the per-partition results are micro-averaged
(i.e., each partition's precision and recall are weighted by its predicted and reference note counts respectively before the F1 is taken). 
Inst F1 is also the faithfulness reward $R_{\text{faith}}$ used during RL (Section~\ref{sec:method:reward}).
\end{itemize}

\paragraph{Readability.}

We report two complementary readability measures, both by LLM judges:
an absolute per-program score (Rubric)
and a relative cross-method score (Rank).

\begin{itemize}[leftmargin=*,itemsep=0.3mm]
\item \textbf{Rubric} is the $12$-item factual checklist score (Section~\ref{sec:method:reward}; Appendix~\ref{app:rubrics}) in $[0,1]$, computed with the same \texttt{Qwen3.6-35B-A3B}~\citep{qwen36_35b_a3b} judge used during RL training. Each program is scored independently; we report the mean rubric across the evaluation set.

\item \textbf{Rank} is the average rank from a listwise LLM reranker~\citep{tang2024found}: 
for each input, the (compiling) outputs of all methods are shown together to an LLM judge that orders them from most to least readable; to control position and label bias we average over multiple random permutations of the candidate order and labels and aggregate the per-permutation orderings into a consensus rank. Lower average rank is better.
\end{itemize}

\paragraph{Diversity.}
In creative settings, 
useful AI suggestions are often 
expected to 
surface multiple plausible
directions rather than 
returning near-identical outputs~\citep{resnick2005design, cococo, kim2025amuse}.
For symbolic music decompilation, 
this means that a model should not collapse to a single programming template: the same musical input can often be expressed through different \Strudel{} idioms, such as chained patterns, shared variables,
layering operators, or section-level structure.
We therefore evaluate diversity as a measure of whether a method explores
different coding idioms across generations.

Among different similarity measurement method~\citep{zhou23codebertscore, bleu, ren2020codebleu} 
we adopt an AST-based metric for programming languages. Pure $n$-gram
measures~\citep{bleu} can be dominated by mini-notation pitches and rhythms,
which obscures idiom-level diversity, whereas embedding-based code-similarity
metrics~\citep{zhou23codebertscore} are trained primarily on high-resource,
mainstream languages and are thus less reliable for a low-resource DSL such as
\Strudel{}. We therefore use the AST-based metric of \citet{ren2020codebleu} for
diversity computation, detailed below.

\begin{itemize}[leftmargin=*,itemsep=0.3mm]
\item \textbf{SelfCB} measures intra-method self-similarity using CodeBLEU~\citep{ren2020codebleu}. 
For a method with programs 
$\{p_1,\dots,p_N\}$ over the shared input set, $\text{CodeBLEU}(\cdot,\cdot)\in[0,1]$ scores code similarity as a weighted combination of standard BLEU, weighted $n$-gram matching, syntactic AST matching, and semantic data-flow matching. We weight the four terms by $(\alpha,\beta,\gamma,\delta)=(0.1,0.1,0.4,0.4)$ which~\citet{ren2020codebleu} find correlate most strongly with human judgments. 
We parse programs with a Acorn JavaScript AST parser~\citep{acorn}. 
As CodeBLEU is asymmetric, we symmetrize each unordered pair and average over all such pairs:

\begin{equation}
  \text{SelfCB}
  = \binom{N}{2}^{-1}\!\!\sum_{i<j}
  \tfrac{1}{2}\!\left(\text{CodeBLEU}(p_i,p_j)+\text{CodeBLEU}(p_j,p_i)\right).
\end{equation}

Analogous to Self-BLEU~\citep{zhu2018texygen}, a lower SelfCB indicates more diverse generations. To compare methods fairly, 
we restrict the pair set to the inputs 
on which all methods compile successfully (and as such exclude vanilla Qwen models which has less than 20\% compile rate to ensure sample size).

\end{itemize}

\subsection{Dataset}
\label{app:eval:dataset}

This section describes the MIDI datasets and sampling method used in our evaluation.
Our experiments cover two settings:
(i) the main evaluation (Section~\ref{sec:exp:main}) and ablation studies (Section~\ref{sec:exp:ablation}) use \dataset{} and short LMD fragments, matching the distributions used during post-training; 
and
(ii) the generalization evaluation (Section~\ref{sec:exp:generalization}) uses additional held-out corpora that introduce distribution shifts in duration, source, and instrumentation.

\paragraph{Main and ablation studies.}
The main experiments (Section~\ref{sec:exp:main}) and ablation studies (Section~\ref{sec:exp:ablation}) 
use two corpora: \dataset{} and LMD. 
\dataset{} is our synthetic paired corpus of executable \Strudel{} programs and rendered MIDI outputs, as described in Section~\ref{sec:method:dataset}.
LMD~\citep{lakhmidi} is a large-scale real-world MIDI corpus of multi-instrument, multi-genre music, 
including arrangements of modern pop songs and transcriptions of Western classical compositions. 
For LMD, we sample short fragments shorter than $30$\,s for evaluation, 
yielding approximately $9$K training segments and $1$K test segments.
Together, \dataset{} and short LMD fragments 
define the in-domain distribution 
against which we evaluate the effects of post-training.

\paragraph{Generalization evaluation.}
To evaluate whether \system{} transfers beyond
the training distributions, we additionally evaluate on four held-out settings. 
For each setting, we sample $200$ evaluation inputs and report Best@$5$ results:
each method generates five candidate programs per input, and we select the candidate with the highest Inst F1.
All remaining metrics are computed on the candidates selected by Inst F1.

\begin{itemize}[leftmargin=*,itemsep=0.3mm]
\item \textbf{LMD}~($30$--$60$\,s; \citealt{lakhmidi}) 
consists of longer LMD excerpts and tests generalization to inputs longer than those used during RL post-training and main evaluation.

\item \textbf{GigaMIDI}~\citep{lee2025gigamidi} 
is a large-scale multi-instrument, multi-genre symbolic music corpus that consolidates several major MIDI resources.
It provides a source-level generalization setting over real-world MIDI files drawn from a broader collection than LMD; 
because GigaMIDI includes LMD among its sources, we exclude LMD-derived files when sampling evaluation inputs.
We use short fragments shorter than $30$\,s.

\item \textbf{NES-MDB}~\citep{donahue2018nesmdb} 
is a multi-instrument symbolic music corpus of approximately $5$K chiptune pieces from $397$ Nintendo Entertainment System games.
Each song is represented with four NES instrument voices, 
corresponding to two pulse channels, one triangle channel, and one noise channel, with annotations for expressive dynamics and timbre attributes.
This setting tests generalization to chiptune music with a source and instrumentation distribution 
that differs substantially from \dataset{} and LMD.
We use short fragments shorter than $30$\,s.

\item \textbf{Nottingham}~\citep{foxley_nottingham} 
is a collection of over $1$K British and American traditional folk tunes, originally represented in ABC notation as lead sheets with monophonic melody and chord accompaniment.
We use the available MIDI versions of the original ABC files.
This setting tests generalization to homophonic lead-sheet music, 
where melodic and chordal roles may be encoded within a compact, 
single-instrument representation rather than separated into distinct instrumental tracks.
Since the pieces are generally longer than the short-fragment setting used for post-training, we crop each piece to a $16$\,s segment, matching the mean duration of the short LMD fragments used during RL post-training.
\end{itemize}

\subsection{Baselines}
\label{app:eval:baseline}

\begin{itemize}[leftmargin=*,itemsep=0.3mm]
\item \textbf{Heuristic converter}~\citep{midi_to_strudel} transliterates MIDI
directly into \Strudel{} mini-notation. 
Starting from the open-source converter, 
which quantizes each note onset to a fixed grid and
emits the nearest \Strudel{} sound, 
we extend it for our setting with: (i)
\emph{drum support}, mapping General MIDI channel-$9$ percussion to \Strudel{}
drum tokens (\eg \texttt{bd}, \texttt{sd}, \texttt{hh}); 
(ii) \emph{multiple-meter support}, 
reading
time-signature changes to 
compute the beats-per-cycle of each segment; 
and 
(iii)
broader \emph{heuristic instrument coverage}, mapping MIDI programs to
\Strudel{} sound names through an expanded lookup table (with a piano fallback
for unmatched melodic programs 
and a synth-drum fallback for percussion). 
This provides a strong reference point for low-level faithfulness but a weak baseline for readable program structure. 

\item \textbf{Frontier LLMs.} 
We compare against three closed-weight 
general-purpose models:
\claude{}~\citep{anthropic2026opus46},
\gemini{}~\citep{google2026gemini35flash},
and \gpt{}~\citep{openai2026gpt55}.
These models represent strong general-purpose code-generation systems
but are not specialized for \Strudel{} or for symbolic music decompilation.
We query each model through its provider's chat API using the same
MIDI-to-text input representation and decompilation prompt described in
Appendix~\ref{app:midi_repr}.
For all models, 
we decode with a maximum of $4096$ new tokens
and disable model-specific extended reasoning modes when available.

\end{itemize}

\subsection{Hyperparameters}
\label{app:eval:hyperparams}
For SFT, we fine-tune both models for one epoch with LoRA~\citep{hu2022lora}
using rank $64$ and $\alpha=128$.
We use a learning rate of $5\times10^{-4}$ with a cosine schedule,
weight decay $0.05$, gradient clipping $1.0$, and batch size $128$.
SFT training is performed on a single node with $8\times$H100 GPUs and completes
in approximately $1.5$ hours.

For RL, we initialize from the SFT checkpoint and optimize with 
rollout group
size $16$, batch size $64$, and learning rate $5\times10^{-6}$ for $10$ epochs.
We use KL coefficient $0.05$, clipping threshold $0.4$, and sampling
temperature $1.0$, a maximum generation length of $4096$ new tokens, and disable the model-specific thinking mode during rollout generation ({\prettyfont enable\_thinking=false}).
We select checkpoints using validation reward on a held-out mixture of \dataset{} and LMD MIDI, 
stopping when the validation faithfulness--readability
tradeoff no longer improves. 
RL training is performed on a single node with $8\times$H100 GPUs and takes
approximately $8$ days.
We implement the training backend using AReaL~\citep{areal}.

\section{Extended Related Work}
\label{app:related}

\paragraph{Multi-objective reinforcement learning.}
RL for language models often optimizes multiple, partially conflicting rewards, such as task accuracy versus reasoning length~\citep{
arora2026training,
aggarwal2025l,
huang2025hapo,
su2025thinking,
xiang2025just,
li2026drpo}
or multiple dimensions of human preference~\citep{dai2024safe, jang2023personalized}. 
A common approach is to 
collapse these objectives into a 
single weighted sum and directly optimizing it~\citep{juan2025rlrfsvg, kolodiazhnyi2026cadrille, aggarwal2025l, arora2026training, huang2025hapo}; yet it often 
removes distinctions across reward dimensions and leads to suboptimal reward convergence.
Recent methods therefore decouple reward signals: DRPO separates the learning signals of correct and incorrect rollouts~\citep{li2026drpo}, while GDPO normalizes each reward component independently before aggregation~\citep{gdpo}. 
Our work adopts GDPO to optimize execution faithfulness and program readability as separate reward components.

\paragraph{Rubric-based evaluation and training.}
Structured rubrics and checklists score LLM outputs along multiple criteria in settings without a single verifiable answer. They are widely used for evaluation~\citep{openai2024healthbench, hashemi2024llmrubric, ruan2026expertlongbench, gandhi26ppteval, gou2025mindweb, pathak2025rubric, dineen2025qalign}, and recent work uses them directly as RL reward signals: Rubrics as Rewards~\citep{gunjal2026rubrics} and RLCF~\citep{viswanathan2025checklists} 
decompose a task into binary checklist items scored by an LLM judge, 
yielding more stable training signal than scalar preference rewards in non-verifiable domains. 
We use the same design for our readability reward, with a fixed checklist combining general code-quality items and Strudel-specific criteria.

\section{Additional Evaluation Results} 
\label{app:more_eval}

This section provides additional results and analysis that complement the experiments in the main paper (Section~\ref{sec:exp}).
We first provide the full generalization results behind the summary in Section~\ref{sec:exp:generalization} (Appendix~\ref{app:more_eval:generalization_full}). 
We then evaluate \system{} across genres to analyze how decompilation faithfulness varies with musical style (Appendix~\ref{app:quantitative:genre}).

\subsection{Full Generalization Results}
\label{app:more_eval:generalization_full}

Table~\ref{tab:apd:generalization_full} reports the full generalization results corresponding to the summary in Table~\ref{tab:generalization} of Section~\ref{sec:exp:generalization}, 
where we evaluate whether \system{} 
generalizes beyond the training distributions. 
The table includes the in-domain \dataset{} and short-fragment LMD settings for reference, 
followed by out-of-domain datasets that vary duration, source, genre, and
instrumentation: longer LMD excerpts~($30$--$60$\,s; \citealt{lakhmidi}), GigaMIDI~\citep{lee2025gigamidi}, NES-MDB~\citep{donahue2018nesmdb}, and Nottingham~\citep{foxley_nottingham}.
For the generalization study, 
we compare \system{} against \gpt{} 
as the representative frontier-LLM baseline. 
Among the frontier models in our main evaluation, \gpt{} provides the strongest overall tradeoff, 
achieving the best Inst F1 and readability rank on both \dataset{} and LMD 
while remaining competitive in diversity (Table~\ref{tab:main}).
We report best@5 results selected by Inst F1; all other metrics are computed on
the selected candidates.

Compared with \gpt{}, \system{} achieves stronger execution faithfulness on most datasets, 
while maintaining comparable readability and diversity. 
The largest gap between \gpt{} and \system{} appears on Nottingham ($+0.25$ Onset F1). 
This may reflect dataset-specific structure 
rather than genre alone: 
Nottingham is a single-instrument, ABC-derived folk collection with comparatively regular symbolic timing,
and our instrument-grouped MIDI representation (Appendix~\ref{app:midi_repr}) collapses melodic and chord accompaniment into the same track. 
Such inputs may make melodic and harmonic organization easier for the learned decompiler to recover.
At the same time, 
although \system{} outperforms \gpt{}, 
absolute faithfulness remains lower on LMD ($30$--$60$\,s) and NES-MDB than in the other settings. 
These cases highlight different aspects of decompilation: 
longer LMD excerpts require structure to be recovered over a wider temporal span, 
while NES-MDB introduces specialized chiptune timbres and dense note patterns. 
For readability and diversity, \system{} obtains higher Rubric scores than \gpt{} across the held-out datasets while Rank and SelfCB remain comparable.
Overall, these results support the main claim of Section~\ref{sec:exp:generalization}: 
the learned decompiler transfers beyond the synthetic paired data, yet its robustness remains sensitive to the temporal and stylistic structure of the input.

\input{tables/generalization_full_table}

\subsection{Genre-wise Faithfulness}
\label{app:quantitative:genre}

To complement the corpus-level generalization study
with a finer-grained analysis of stylistic variation,
we evaluate \system{} on ModArchive,\footnote{\prettyurl{https://modarchive.org}}
a public archive of tracker-format music.
A tracker module is a self-contained music file
that stores
symbolic note patterns
together with instrument samples and sequencing information.
By combining sample-level audio material with explicit symbolic structure,
tracker modules provide a natural intermediate representation 
for the broader audio-to-code decompilation setting (Section~\ref{sec:exp:application}).
ModArchive is also well matched to our evaluation because
tracker music and \Strudel{} are both closely tied to 
electronic and sample-based music-making practices,
and many modules include community-provided genre tags 
which allow us to analyze decompilation faithfulness across fine-grained musical styles.

We crawl tracker modules from ModArchive across $51$ genre tags, 
recording each module's genre, format, and rating.
For each genre, we target the top $100$ modules by rating. 
Because tracker formats (\eg {\prettyfont .mod}, {\prettyfont .xm}, {\prettyfont .it}, {\prettyfont .s3m})
differ substantially from the General MIDI inputs used elsewhere in this paper,
we convert each module to GM MIDI using a headless wrapper around the open-source MilkyTracker engine.\footnote{\prettyurl{https://milkytracker.org/}} 
Each tracker instrument is mapped to a General MIDI program using \texttt{Qwen2.5-Omni}~\citep{Qwen2.5-Omni}; 
samples without a clear pitched or percussive identity, such as speech or sound effects, are discarded.
We then extract a $16$\,s segment from each valid module,
matching the mean duration of the short LMD fragments used during RL post-training (Appendix~\ref{app:eval}). 
The resulting benchmark contains $4{,}463$ inputs across all $51$ genres, 
with an average of $87.5$ examples per genre (range: $61$--$97$).\footnote{The final number of examples per genre is below the initial crawl target of $100$ because some genres contain fewer than $100$ modules, or because some modules become empty after MIDI conversion. See \prettyurl{https://modarchive.org/index.php?request=view_genres} for the genre categories and their listed module counts.}
As in the generalization study, 
we evaluate \system{}-8B using Best@$5$ metrics (Section~\ref{sec:exp:generalization}).

\input{tables/modarchive_genre_table}

Table~\ref{tab:apd:genre} and Figure~\ref{fig:apd:genre} 
report per-genre faithfulness, sorted by Onset F1.
Performance varies substantially across genres, 
while compile rates remain almost saturated: 
all genres achieve a compile rate of $1.00$ except Electronic--Gabber ($0.99$) and Classical ($0.98$).
The genre-wise gaps therefore 
reflect differences in musical reconstruction rather 
than failures to produce executable \Strudel{} programs.

At the level of broad genre families,
Table~\ref{tab:apd:genre} shows that 
Electronica achieves the highest aggregated scores, 
followed by Seasonal, Alternative, and Pop/Rock.
The fine-grained ranking in Figure~\ref{fig:apd:genre} shows a similar trend:
dance- and electronic-oriented genres tend to score higher,
with Disco ranked first and 
Electronica genres such as Jungle, Minimal, Acid, and House concentrated near the top. 
This pattern is consistent with the close fit between \Strudel{}'s pattern-based abstractions
and rhythmically regular electronic and dance music.
By contrast, lower-scoring genres include 
Jazz, Funk, Chiptune, Orchestral, and Reggae.
These genres often involve expressive timing, 
off-beat rhythmic patterns such as syncopation, 
or dense note and instrumentation patterns, 
which can make the input harder to recover 
as a concise executable \Strudel{} program.
Overall, the results suggest that \system{} 
transfers best to genres whose structure is well captured by repeated pattern abstractions, 
while highlighting rhythmic variation and dense textures as key directions for improving faithful decompilation.

\input{figures/fig_apd_genre}

\section{Qualitative Examples}
\label{app:qualitative}
Figures~\ref{fig:apd:qual_examples_1} and~\ref{fig:apd:qual_examples_2} show qualitative examples comparing decompilations of the same MIDI input across representative methods: 
\system{} (8B), the vanilla Qwen3-8B model (\texttt{Qwen3-8B}; \citealt{qwen3technicalreport}), 
Qwen3-8B after SFT, 
the heuristic converter~\citep{midi_to_strudel}, 
and the three frontier models used in our evaluation:  \claude{}~\citep{anthropic2026opus46}, \gemini{}~\citep{google2026gemini35flash}, and \gpt{}~\citep{openai2026gpt55}.
To make structural differences easier to compare, we omit sound effect parameters such as {\prettyfont gain()} and {\prettyfont room()}.
For additional examples with sound, please refer to the project page: \prettyurl{https://yewon-kim.com/decomposer}.

\input{figures/fig_apd_qual_examples_1}
\input{figures/fig_apd_qual_examples_2}

%% file: figures/fig_apd_strudel_examples.tex
\begin{figure*}[t!]
    \centering
    \begin{subfigure}[t]{0.49\linewidth}
        \centering
        \includegraphics[width=\linewidth]{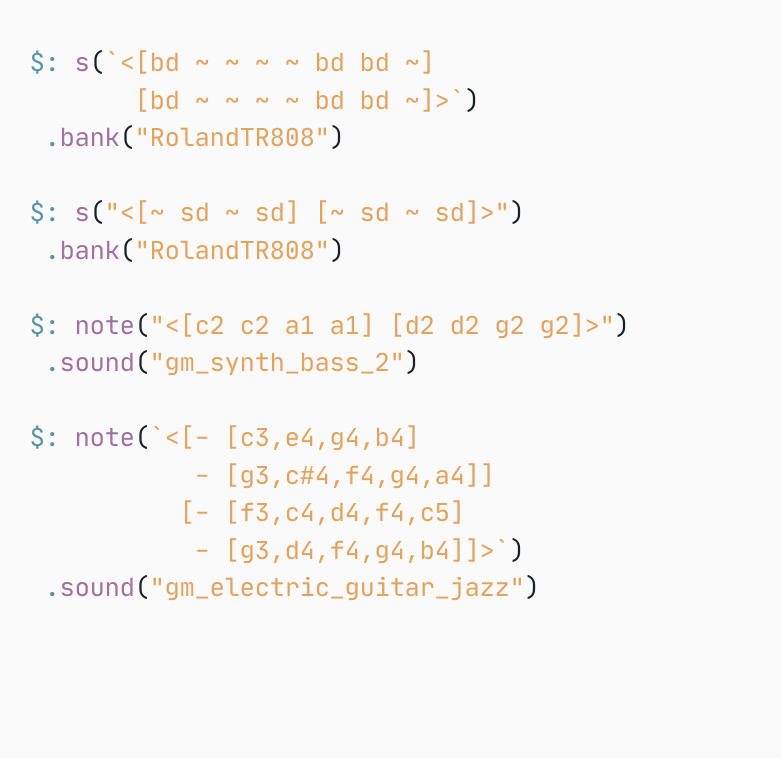}
        \caption{\textbf{Event-level transliteration.} 
        Musical contents are quantized into mini-notation strings that explicitly
        list drum hits, pitches, and rests.}
        \label{fig:apd:strudel:literal}
    \end{subfigure}
    \hfill
    \begin{subfigure}[t]{0.49\linewidth}
        \centering
        \includegraphics[width=\linewidth]{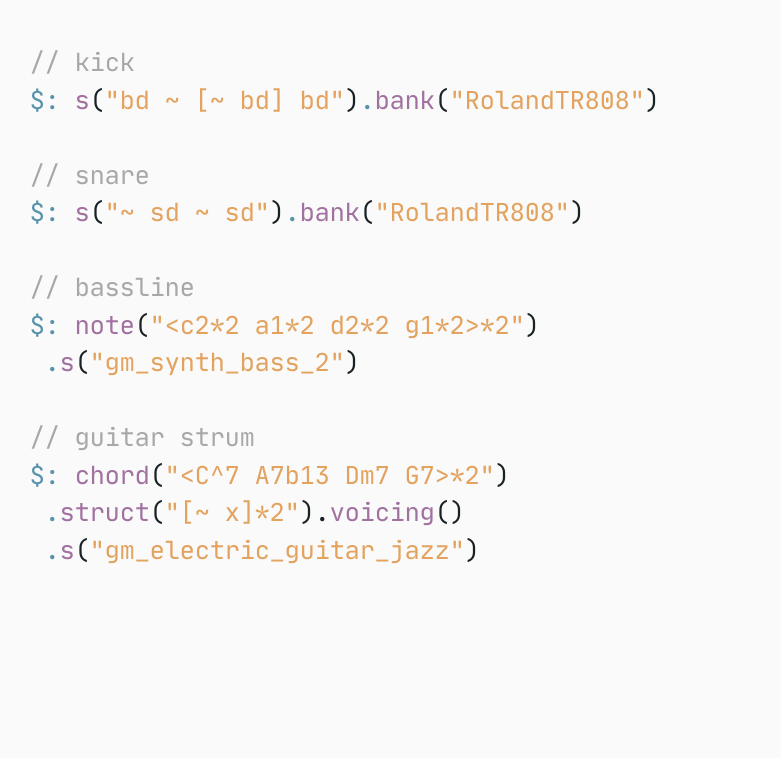}
        \caption{\textbf{Pattern-level abstraction.} 
        The bassline compacts events with pattern operators, 
        and the guitar layer represents harmony with {\prettyfont chord()} and rhythm with {\prettyfont struct()}. 
}
        \label{fig:apd:strudel:mini}
    \end{subfigure}

    \vspace{0.5em}

    \begin{subfigure}[t]{0.49\linewidth}
        \centering
        \includegraphics[width=\linewidth]{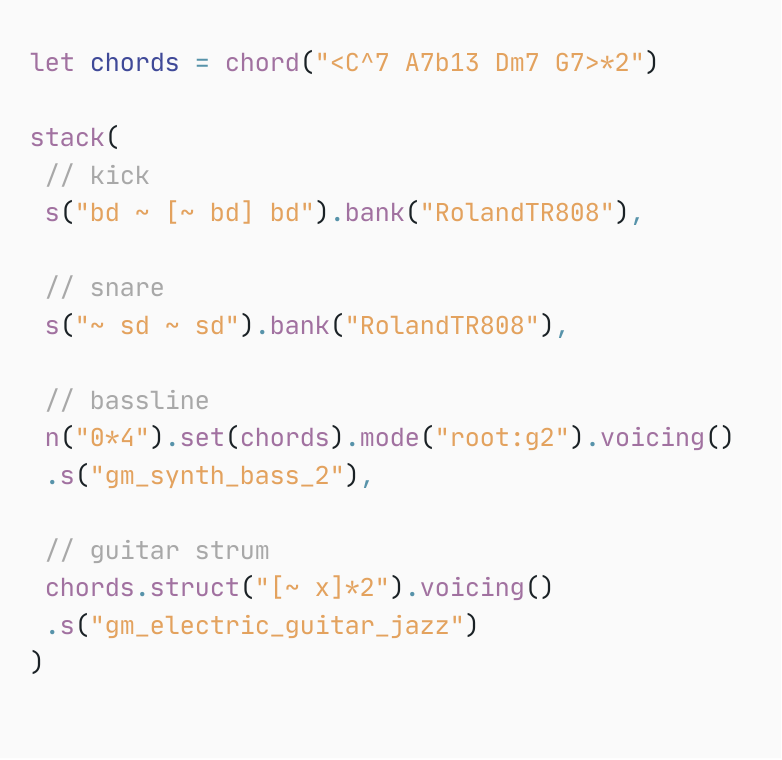}
        \caption{\textbf{Shared structure and layering.} 
        A shared {\prettyfont chords} variable anchors the bassline and the guitar harmony, 
        while {\prettyfont stack()} groups the layers as simultaneous voices.
        }
        \label{fig:apd:strudel:stack}
    \end{subfigure}
    \hfill
    \begin{subfigure}[t]{0.49\linewidth}
        \centering
        \includegraphics[width=\linewidth]{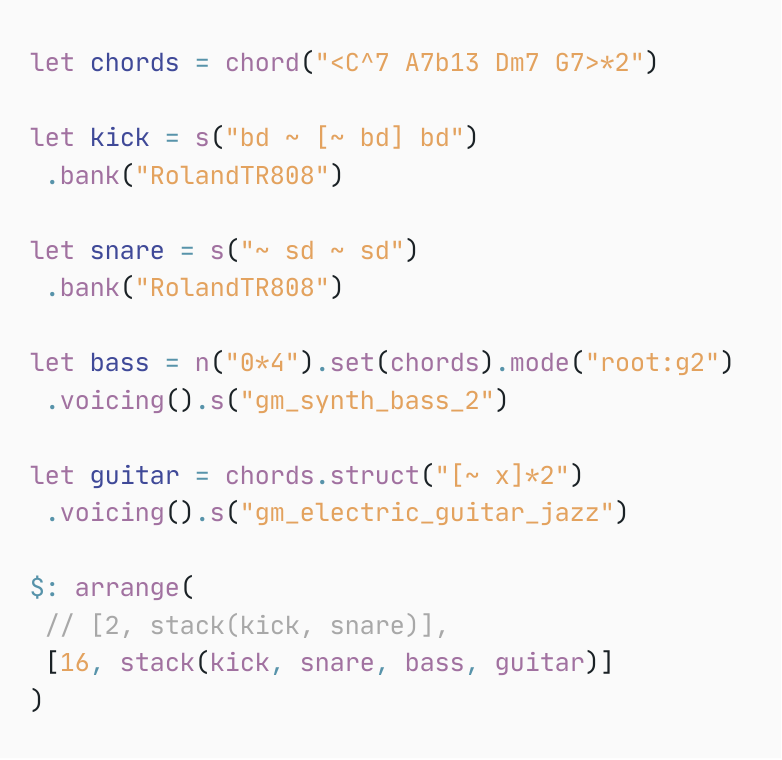}
        \caption{\textbf{Arrangement-level organization.} 
        Voices are factored into named variables, and their combinations are sequenced into sections with 
        {\prettyfont arrange()}.}
        \label{fig:apd:strudel:arrange}
    \end{subfigure}

    \caption{\textbf{Different \Strudel{} representations of the same musical loop.} 
    The same musical content can be expressed by 
    many distinct \Strudel{} programs.
    Among many possible programs for the same musical content, 
    we show four illustrative examples that render to the 
    same short loop of drums, bass, and electric guitar.
    They differ in their levels of abstraction and coding idioms, ranging from literal event-level representation to pattern-, layer-, and arrangement-level abstractions.
    }
    \vspace{-1.5em}
    \label{fig:apd:strudel_examples}
\end{figure*}

%% file: tables/regex_table.tex
\begin{table}[t]
\centering
\caption{\textbf{Deterministic regex rewrites.}
Regex-based fixes applied to LLM-generated \Strudel{} programs during data cleaning.
Each rule targets a recurring syntactic mistake 
observed in initial generations generated by \claude{}; 
together, they correct ${\sim}37\%$ of generated programs.
}
\label{tab:apd:regex_fixes}
\small
\setlength{\tabcolsep}{5pt}
\renewcommand{\arraystretch}{1.32}
\newcommand{\inentry}[1]{%
  {\renewcommand{\arraystretch}{1}%
   \begin{tabular}[t]{@{}l@{}}#1\end{tabular}}}
\begin{tabularx}{\linewidth}{@{}>{\RaggedRight\arraybackslash}p{0.34\linewidth}
                              >{\RaggedRight\arraybackslash}X@{}}
\toprule
\textbf{Rewrite} & \textbf{Description} \\
\midrule
\rowcolor{glightpurple}[1pt][0pt]
\multicolumn{2}{@{}l@{}}{\hspace{0.5em}\textbf{Invalid sound aliases}} \\
{\prettyfont gm\_electric\_piano\_1}
\(\to\)
{\prettyfont gm\_epiano1}
&
Correct the invalid e-piano alias to the valid \Strudel{} sound name. \\
{\prettyfont \textbackslash bshaker\textbackslash b(?!\_)}
\(\to\)
{\prettyfont sh}
&
Replace the invalid bare {\prettyfont shaker} alias with {\prettyfont sh};
keep {\prettyfont shaker\_small} and {\prettyfont shaker\_large} as-is. \\
\addlinespace[2pt]
\rowcolor{glightpurple}[1pt][0pt]
\multicolumn{2}{@{}l@{}}{\hspace{0.5em}\textbf{Voicing misuse}} \\
{\prettyfont \textbackslash.voicings\textbackslash(".*?"\textbackslash)}
\(\to\)
{\prettyfont .voicing()}
&
Collapse {\prettyfont .voicings(arg)} calls to 
{\prettyfont .voicing()}. \\
\inentry{%
{\prettyfont 
\textbackslash.note\textbackslash((".*?"\textbar[\textbackslash w\$]+)\textbackslash.voicing}
\\
{\prettyfont \textbackslash(\textbackslash)\textbackslash)} 
\(\to\) {\prettyfont .chord(\$1).voicing()}%
}
&
Repair {\prettyfont note(X.voicing())} by rewriting it as a
{\prettyfont chord().voicing()} call. \\
\inentry{%
{\prettyfont \textbackslash b(?:note|n)\textbackslash s*\textbackslash((.*?)\textbackslash)} \\
{\prettyfont (?=\textbackslash s*\textbackslash.voicing\textbackslash(\textbackslash))}
\(\to\)
{\prettyfont chord(\$1)}%
}
&
Rewrite voiced {\prettyfont note()} / {\prettyfont n()} heads as {\prettyfont chord()}. \\
\addlinespace[2pt]
\rowcolor{glightpurple}[1pt][0pt]
\multicolumn{2}{@{}l@{}}{\hspace{0.5em}\textbf{Erroneous note calls}} \\
{\prettyfont \textbackslash.note\textbackslash(x => [\textasciicircum)]*\textbackslash)}
\(\to\)
\(\emptyset\)
&
Remove lambda-style {\prettyfont .note()} calls. \\
\inentry{%
{\prettyfont\detokenize{\.(rootNotes|voicing|scale)}} \\
{\prettyfont\detokenize{\([^)]*\).note()}} 
\(\to\) remove {\prettyfont .note()}%
}
&
Remove empty {\prettyfont .note()} calls after tonal operators. \\
{\prettyfont stack(}\(\dots\){\prettyfont ).note()}
\(\to\)
{\prettyfont stack(}\(\dots\){\prettyfont )}
&
Remove a {\prettyfont .note()} appended to the 
{\prettyfont stack(}\(\dots\){\prettyfont )} call. \\
\addlinespace[2pt]
\rowcolor{glightpurple}[1pt][0pt]
\multicolumn{2}{@{}l@{}}{\hspace{0.5em}\textbf{Broken chain structure}} \\
{\prettyfont rootNotes} restructure
&
Reorder {\prettyfont s(sample).note(expr.rootNotes(N))} to
{\prettyfont expr.s(sample)}, correcting broken bass-line chains. \\
\addlinespace[2pt]
\rowcolor{glightpurple}[1pt][0pt]
\multicolumn{2}{@{}l@{}}{\hspace{0.5em}\textbf{Length and trigger guards}} \\
{\prettyfont arrange()} section lengths
&
Cap each {\prettyfont arrange()} section at \(\leq 2\) cycles and the total arrangement
at \(\leq 12\) cycles to prevent overly long outputs. \\
\inentry{%
{\prettyfont .struct("0 0")} 
\(\to\)
{\prettyfont .struct("1 0")}%
}
&
When a {\prettyfont .struct(}\(\dots\){\prettyfont )} pattern contains no trigger character
({\prettyfont 1}, {\prettyfont t}, or {\prettyfont x}), 
replace the first {\prettyfont 0} with {\prettyfont 1}. \\
\bottomrule
\end{tabularx}
\vspace{-1em}
\end{table}

%% file: figures/fig_apd_data_cleaning.tex
\begin{figure*}[t!]
    \centering
    \includegraphics[width=\linewidth]{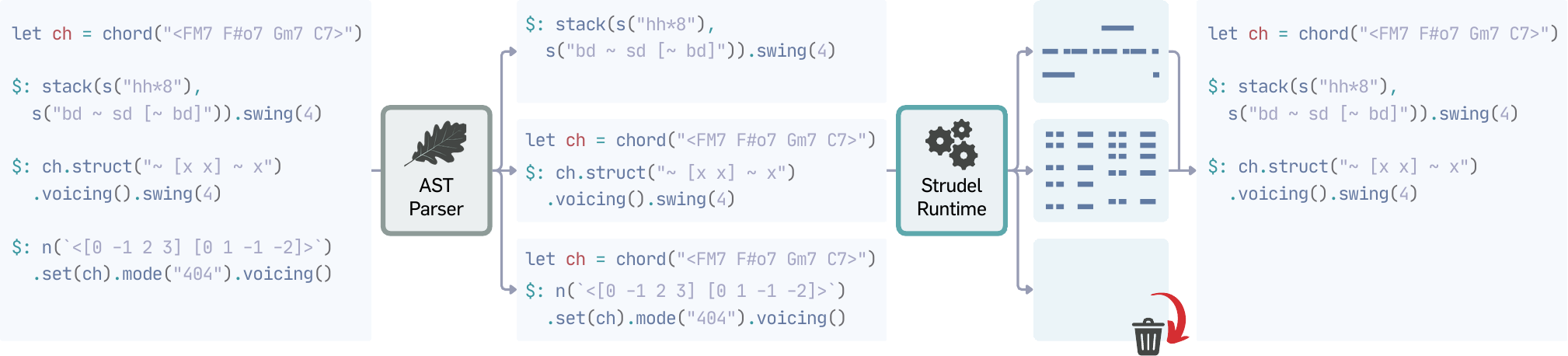}
    \caption{
    \textbf{Voice-level cleaning pipeline.} 
    After regex-level cleaning of LLM-generated programs, we use an AST parser to extract each top-level musical voice along with the declarations and configuration calls it depends on.
    Each voice is rendered in isolation 
    and removed if it produces no note-on events.
    We then remove unused declarations and keep programs with at least one surviving voice.
    }
    \vspace{-1em}
    \label{fig:apd:data_cleaning}
\end{figure*}

%% file: tables/music_seed_table.tex

\definecolor{everynoise}{HTML}{949490}
\definecolor{theorytab}{HTML}{c7bd99}
\definecolor{midicaps}{HTML}{6db0ae}

\newcommand{\attrEN}[1]{\textcolor{everynoise}{{\prettyfont\normalsize #1}}}
\newcommand{\attrTT}[1]{\textcolor{theorytab}{{\prettyfont\normalsize #1}}}
\newcommand{\attrMC}[1]{\textcolor{midicaps}{{\prettyfont\normalsize #1}}}

\newcommand{\seedentry}[1]{%
  {\renewcommand{\arraystretch}{1}%
   \begin{tabular}[t]{@{}l@{}}#1\end{tabular}}}

\begin{wraptable}{r}{0.48\linewidth}
\vspace{-0.95em}
\centering
\footnotesize
\caption{
    \textbf{Musical conditioning seeds.}
    The conditioning modes 
    range from genre-only prompts to 
    richer multi-attribute prompts.
    Text color indicates the source of each exposed musical attribute:
    \textcolor{everynoise}{EveryNoise},
    \textcolor{theorytab}{TheoryTab}, and
    \textcolor{midicaps}{MidiCaps}.
}
\setlength{\tabcolsep}{3.6pt}
\renewcommand{\arraystretch}{1.2}
\begin{tabularx}{\linewidth}{@{}c
    >{\RaggedRight\arraybackslash}p{0.24\linewidth}
    >{\RaggedRight\arraybackslash}X@{}}
\toprule
\textbf{\# Attrs} & \textbf{Source} & \textbf{Music Attributes} \\
\midrule

$1$ &
\textcolor{everynoise}{EveryNoise} &
\attrEN{genre} \\[5pt]

$5$ &
\seedentry{%
\textcolor{everynoise}{EveryNoise}\\
\textcolor{theorytab}{TheoryTab}} &
\seedentry{
\attrEN{genre},
\attrTT{key},
\attrTT{tempo} \attrTT{(BPM)},\\
\attrTT{beats per bar},\\
\attrTT{chord progression}} \\[5pt]

$7$ &
\textcolor{midicaps}{MidiCaps} &
\seedentry{
\attrMC{genre},
\attrMC{key},
\attrMC{mood}, 
\attrMC{tempo} \\
\attrMC{(BPM)},
\attrMC{beats per bar},\\
\attrMC{instrumentation},\\
\attrMC{chord progression}} \\

\bottomrule
\end{tabularx}
\label{tab:apd:music_seed}
\vspace{-0.4em}
\vspace{-0.9em}
\end{wraptable}

%% file: tables/generalization_full_table.tex
\begin{table*}[t]
\centering
\small
\setlength{\tabcolsep}{4pt}
\renewcommand{\arraystretch}{1.10}
\caption{
\textbf{Full generalization results across diverse MIDI datasets.} 
We compare \gpt{} and \system{} (8B) on in-domain and out-of-domain datasets,
reporting Best@5 results selected by Inst F1.
\system{} achieves stronger overall faithfulness scores than \gpt{} across datasets,
while maintaining comparable readability and diversity.
$\uparrow$/$\downarrow$ indicate whether higher/lower is better;
\textbf{bold} marks the better result between the two methods, and
\underline{underline} marks a tie.
}
\begin{tabular}{llccccccc}
\toprule
&& \multicolumn{4}{c}{\textbf{Faithfulness}}
& \multicolumn{2}{c}{\textbf{Readability}}
& \textbf{Diversity} \\
\cmidrule(lr){3-6}
\cmidrule(lr){7-8}
\cmidrule(lr){9-9}
\textbf{Dataset} & \textbf{Method}
& $\uparrow$Comp.
& $\uparrow$Onset
& $\uparrow$Frame
& $\uparrow$Inst
& $\uparrow$Rubric
& $\downarrow$Rank
& $\downarrow$SelfCB \\
\midrule

\rowcolor{glightpurple}
\multicolumn{9}{l}{\hspace{0.5em}\textbf{In-domain evaluation}} \\
\addlinespace[2pt]

\multirow{2}{*}{\makecell[l]{\dataset{}}}
& \gpt{}     & 0.97 & 0.70 & 0.67 & 0.62 & 0.37 & 1.61 & \textbf{0.45} \\
& \system{} & \textbf{1.00} & \textbf{0.86} & \textbf{0.74} & \textbf{0.78} & \textbf{0.69} & \textbf{1.39} & 0.47 \\
\addlinespace[2pt]

\multirow{2}{*}{LMD ($<30$\,s)}
& \gpt{}     & 0.94 & 0.44 & 0.41 & 0.39 & 0.29 & \textbf{1.44} & 0.54 \\
& \system{} & \textbf{1.00} & \textbf{0.70} & \textbf{0.57} & \textbf{0.68} & \textbf{0.56} & 1.56 & \textbf{0.46} \\

\addlinespace[2pt]
\midrule
\addlinespace[2.5pt]

\rowcolor{glightpurple}
\multicolumn{9}{l}{\hspace{0.5em}\textbf{Out-of-domain evaluation}} \\
\addlinespace[2pt]

\multirow{2}{*}{LMD ($30$--$60$\,s)}
& \gpt{}     & 0.83 & 0.21 & 0.25 & 0.13 & 0.31 & 1.81 & 0.43 \\
& \system{} & \textbf{1.00} & \textbf{0.35} & \textbf{0.31} & \textbf{0.31} & \textbf{0.68} & \textbf{1.19} & \textbf{0.42} \\
\addlinespace[2pt]

\multirow{2}{*}{GigaMIDI}
& \gpt{}     & 0.99 & 0.57 & \textbf{0.38} & 0.57 & 0.25 & 1.61 & 0.55 \\
& \system{} & \textbf{1.00} & \textbf{0.61} & 0.35 & \textbf{0.61} & \textbf{0.56} & \textbf{1.39} & \textbf{0.47} \\
\addlinespace[2pt]

\multirow{2}{*}{NES-MDB}
& \gpt{}     & 0.91 & 0.37 & 0.33 & 0.28 & 0.28 & \textbf{1.49} & 0.49 \\
& \system{} & \textbf{1.00} & \textbf{0.43} & \textbf{0.35} & \textbf{0.40} & \textbf{0.61} & 1.51 & \textbf{0.48} \\
\addlinespace[2pt]

\multirow{2}{*}{Nottingham}
& \gpt{}     & \underline{1.00} & 0.51 & 0.46 & 0.51 & 0.28 & \underline{1.50} & \textbf{0.46} \\
& \system{} & \underline{1.00} & \textbf{0.76} & \textbf{0.65} & \textbf{0.76} & \textbf{0.45} & \underline{1.50} & 0.49 \\

\bottomrule
\end{tabular}
\label{tab:apd:generalization_full}
\end{table*}

%% file: tables/modarchive_genre_table.tex
\colorlet{GroupHighlight}{glightpurple}
\newcommand{\gcell}[1]{\cellcolor{GroupHighlight}\textbf{#1}}
\newcommand{\grouprow}[5]{%
  \gcell{#1} & \gcell{#2} & \gcell{#3} & \gcell{#4} & \gcell{#5}%
}

\newcommand{\subgenre}[1]{\hspace{0.7em}#1}


\begin{table*}[t]
\centering
\scriptsize
\caption{
\textbf{Per-genre faithfulness on ModArchive.}
We report Best@5 faithfulness of \system{} (8B) across 51 ModArchive genres,
grouped by ModArchive's top-level genre families. Group rows report sample-weighted averages.
Compile rate is omitted because it is $1.00$ for all genres except Electronic -- Gabber ($0.99$)
and Classical ($0.98$).
}
\label{tab:apd:genre}
\setlength{\tabcolsep}{3.2pt}
\renewcommand{\arraystretch}{1.03}
\begin{adjustbox}{width=0.98\textwidth,center}
\begin{tabular}{@{\hspace{0.5em}}
  >{\raggedright\arraybackslash}p{0.245\linewidth} r c c c
  @{\hspace{1.25em}}
  >{\raggedright\arraybackslash}p{0.245\linewidth} r c c c
@{\hspace{0.5em}}}
\toprule
Genre & $n$ & Onset & Frame & Inst
&
Genre & $n$ & Onset & Frame & Inst \\
\midrule

\textbf{\textit{Overall}} &
\textbf{4463} & \textbf{0.38} & \textbf{0.37} & \textbf{0.37}
&
\hspace{1em}-- & -- & -- & -- & -- \\ 
\midrule

\grouprow{Electronica}{1763}{0.40}{0.38}{0.39}
&
\grouprow{Pop / Rock}{925}{0.38}{0.37}{0.37} \\

\subgenre{Electronic -- Jungle} & 93 & 0.47 & 0.38 & 0.46
&
\subgenre{Disco} & 97 & 0.48 & 0.42 & 0.46 \\

\subgenre{Electronic -- Minimal} & 66 & 0.46 & 0.37 & 0.45
&
\subgenre{Ballad} & 92 & 0.42 & 0.35 & 0.40 \\

\subgenre{Trance -- Acid} & 93 & 0.44 & 0.41 & 0.41
&
\subgenre{Pop -- Synth} & 82 & 0.41 & 0.41 & 0.39 \\

\subgenre{Trance -- Dream} & 84 & 0.44 & 0.38 & 0.42
&
\subgenre{Pop -- Soft} & 97 & 0.41 & 0.36 & 0.39 \\

\subgenre{Electronic (general)} & 91 & 0.43 & 0.42 & 0.42
&
\subgenre{Rock (general)} & 92 & 0.37 & 0.37 & 0.36 \\

\subgenre{Electronic -- House} & 92 & 0.42 & 0.43 & 0.41
&
\subgenre{Pop (general)} & 93 & 0.37 & 0.35 & 0.35 \\

\subgenre{Trance (general)} & 77 & 0.42 & 0.40 & 0.40
&
\subgenre{Easy Listening} & 95 & 0.36 & 0.36 & 0.34 \\

\subgenre{Trance -- Goa} & 71 & 0.41 & 0.37 & 0.38
&
\subgenre{Rock -- Hard} & 91 & 0.36 & 0.38 & 0.35 \\

\subgenre{Electronic -- Industrial} & 86 & 0.40 & 0.40 & 0.39
&
\subgenre{Rock -- Soft} & 94 & 0.34 & 0.33 & 0.34 \\

\subgenre{Electronic -- Gabber} & 90 & 0.40 & 0.40 & 0.40
&
\subgenre{Funk} & 92 & 0.31 & 0.36 & 0.29 \\

\subgenre{Trance -- Hard} & 61 & 0.40 & 0.38 & 0.38
&
\grouprow{Other}{978}{0.37}{0.34}{0.35} \\

\subgenre{Electronic -- Rave} & 92 & 0.40 & 0.39 & 0.38
&
\subgenre{New Age} & 89 & 0.41 & 0.37 & 0.40 \\

\subgenre{Electronic -- Progressive} & 80 & 0.39 & 0.38 & 0.38
&
\subgenre{Experimental} & 92 & 0.40 & 0.37 & 0.40 \\

\subgenre{Electronic -- Drum \& Bass} & 85 & 0.39 & 0.38 & 0.38
&
\subgenre{Piano} & 94 & 0.38 & 0.30 & 0.37 \\

\subgenre{Electronic -- Dance} & 89 & 0.38 & 0.40 & 0.37
&
\subgenre{Comedy} & 94 & 0.37 & 0.37 & 0.36 \\

\subgenre{Electronic -- IDM} & 77 & 0.38 & 0.37 & 0.38
&
\subgenre{Video Game} & 92 & 0.37 & 0.37 & 0.35 \\

\subgenre{Electronic -- Breakbeat} & 88 & 0.38 & 0.38 & 0.37
&
\subgenre{Soundtrack} & 79 & 0.36 & 0.33 & 0.35 \\

\subgenre{Electronic -- Techno} & 90 & 0.37 & 0.37 & 0.35
&
\subgenre{Fantasy} & 86 & 0.36 & 0.35 & 0.35 \\

\subgenre{Electronic -- Ambient} & 86 & 0.37 & 0.38 & 0.36
&
\subgenre{Folk} & 91 & 0.36 & 0.34 & 0.35 \\

\subgenre{Chillout} & 89 & 0.35 & 0.32 & 0.33
&
\subgenre{World} & 91 & 0.34 & 0.32 & 0.32 \\

\subgenre{Electronic -- Hardcore} & 83 & 0.33 & 0.33 & 0.32
&
\subgenre{Classical} & 88 & 0.33 & 0.30 & 0.32 \\

& & & & 
&
\subgenre{Orchestral} & 82 & 0.32 & 0.30 & 0.31 \\

\addlinespace[0.2em]

\grouprow{Seasonal}{93}{0.39}{0.36}{0.38}
&
\grouprow{Urban / Hip-Hop}{167}{0.36}{0.38}{0.33} \\

\subgenre{Christmas} & 93 & 0.39 & 0.36 & 0.38
&
\subgenre{Hip-Hop} & 91 & 0.39 & 0.38 & 0.36 \\

& & & &
&
\subgenre{Reggae} & 76 & 0.33 & 0.38 & 0.31 \\

\addlinespace[0.2em]

\grouprow{Alternative}{185}{0.39}{0.38}{0.37}
&
\grouprow{Jazz / Blues}{255}{0.32}{0.36}{0.31} \\

\subgenre{Alternative} & 94 & 0.39 & 0.39 & 0.38
&
\subgenre{Jazz -- Acid} & 67 & 0.34 & 0.39 & 0.33 \\

\subgenre{Metal (general)} & 91 & 0.38 & 0.38 & 0.36
&
\subgenre{Jazz -- Modern} & 93 & 0.34 & 0.35 & 0.32 \\

& & & &
&
\subgenre{Jazz (general)} & 95 & 0.30 & 0.35 & 0.28 \\

\addlinespace[0.2em]

& & & &
&
\grouprow{Demo / Tracking Scene}{97}{0.31}{0.34}{0.30} \\

& & & &
&
\subgenre{Chiptune} & 97 & 0.31 & 0.34 & 0.30 \\

\bottomrule
\end{tabular}
\end{adjustbox}
\end{table*}

%% file: figures/fig_apd_genre.tex
\begin{figure*}[t!]
    \centering
    \includegraphics[width=\linewidth]{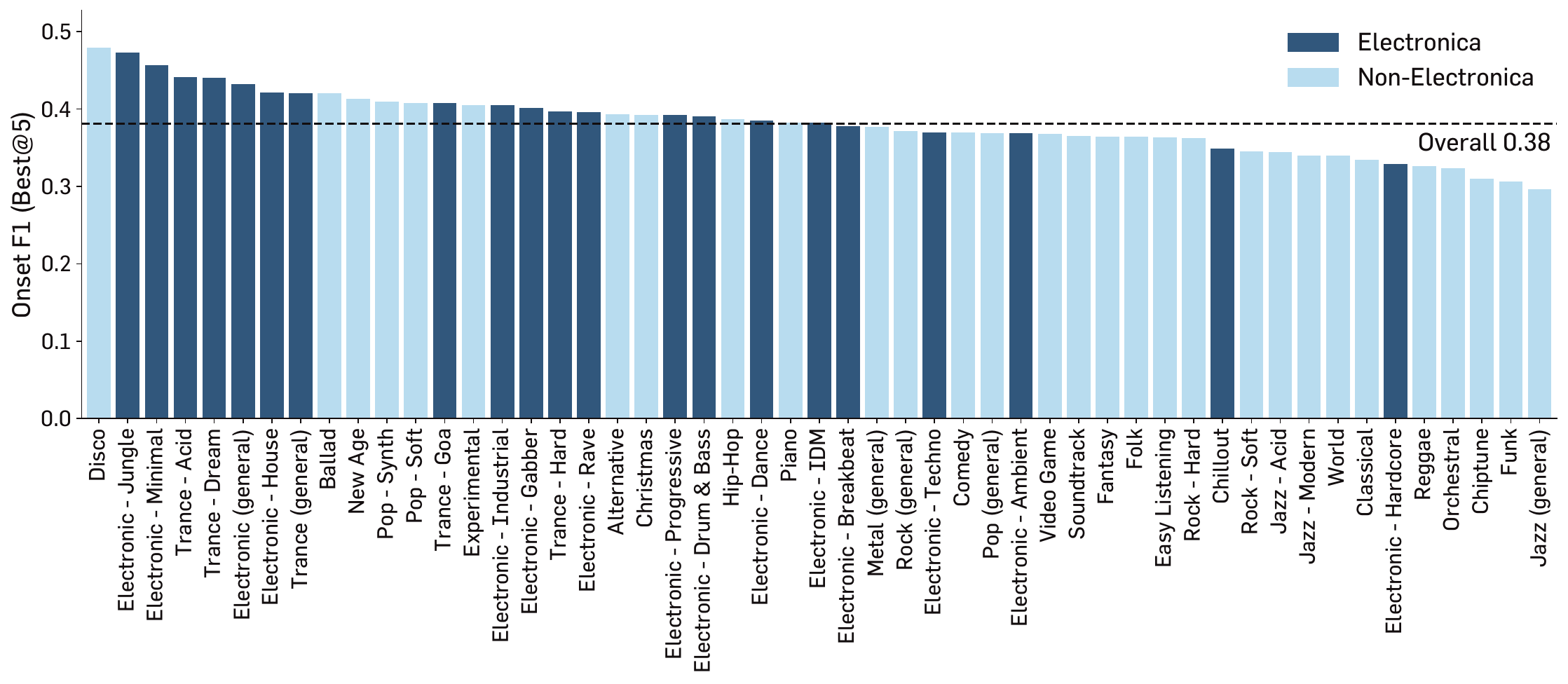}
    \caption{ 
    \textbf{Genre-wise faithfulness on ModArchive.} 
    We show Best@$5$ Onset F1 of \system{} (8B) across $51$ ModArchive genres, sorted in descending order. Electronica genres are highlighted separately from other genres, and the dashed line indicates the overall average. 
    The distribution shows that performance varies substantially across musical styles, 
    with several electronic and dance--oriented genres above average
    and
    genres with more varied instrumentation or less regular rhythmic structure more often appearing near the lower end.
    }
    \label{fig:apd:genre}
\end{figure*}

%% file: figures/fig_apd_qual_examples_1.tex
\begin{figure*}[t!]
    \centering
    \begin{subfigure}[t]{0.49\linewidth}
        \centering
        \includegraphics[width=\linewidth]{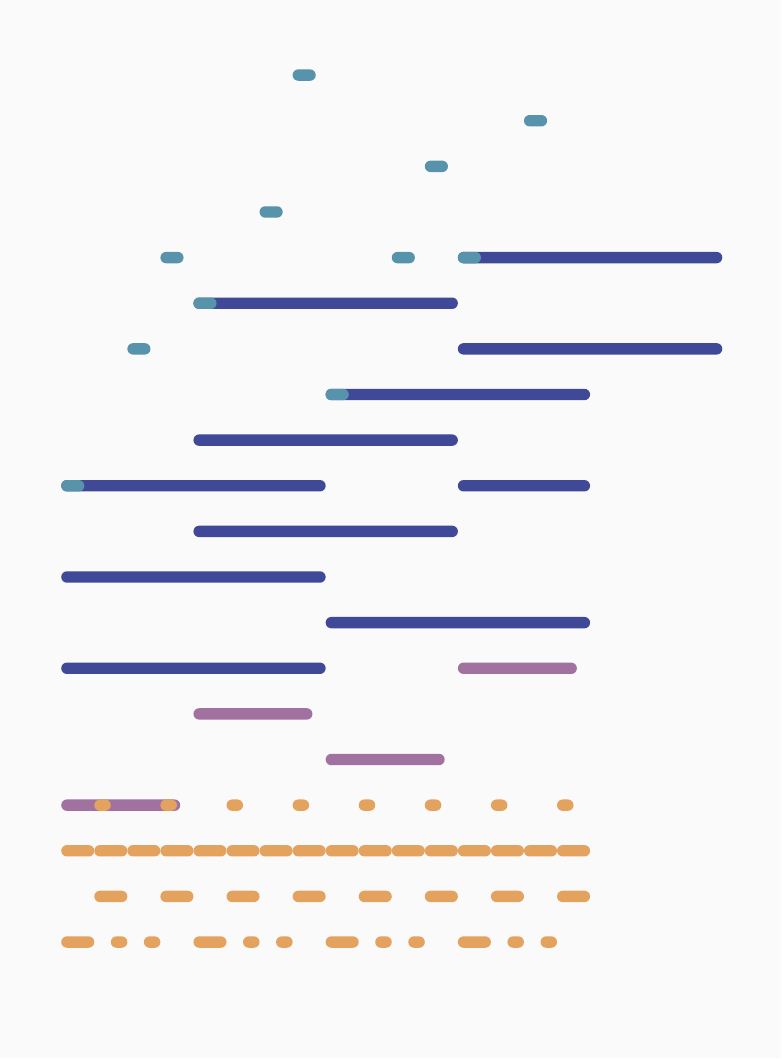}
        \caption{Piano roll visualization of the input MIDI}
        \label{fig:apd:qual:input}
    \end{subfigure}
    \hfill
    \begin{subfigure}[t]{0.49\linewidth}
        \centering
        \includegraphics[width=\linewidth]{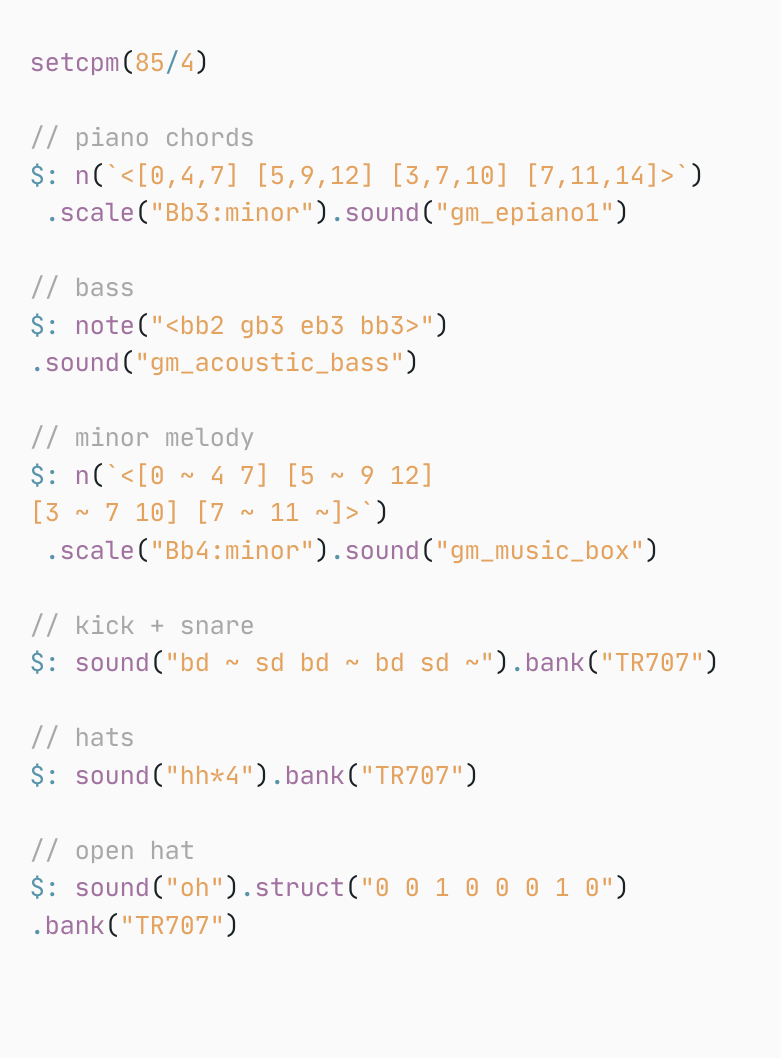}
        \caption{\system{} (8B)}
        \label{fig:apd:qual:decomposer}
    \end{subfigure}

    \vspace{0.5em}

    \begin{subfigure}[t]{0.49\linewidth}
        \centering
        \includegraphics[width=\linewidth]{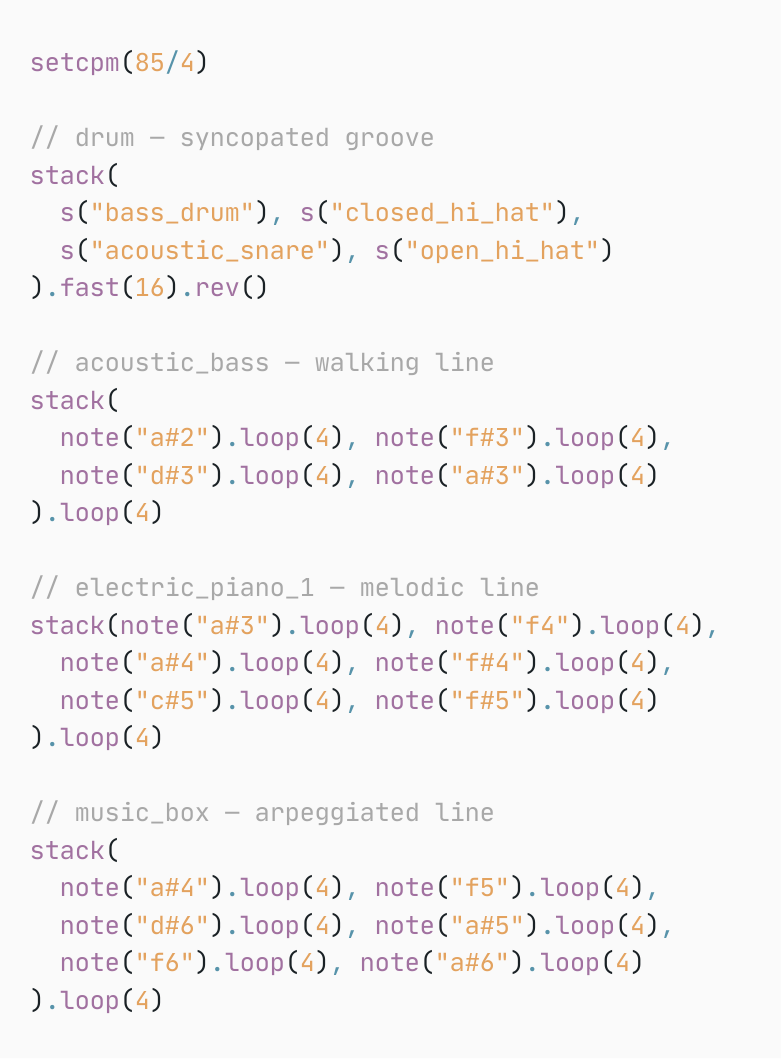}
        \caption{Vanilla Qwen3-8B}
        \label{fig:apd:qual:vanilla}
    \end{subfigure}
    \hfill
    \begin{subfigure}[t]{0.49\linewidth}
        \centering
        \includegraphics[width=\linewidth]{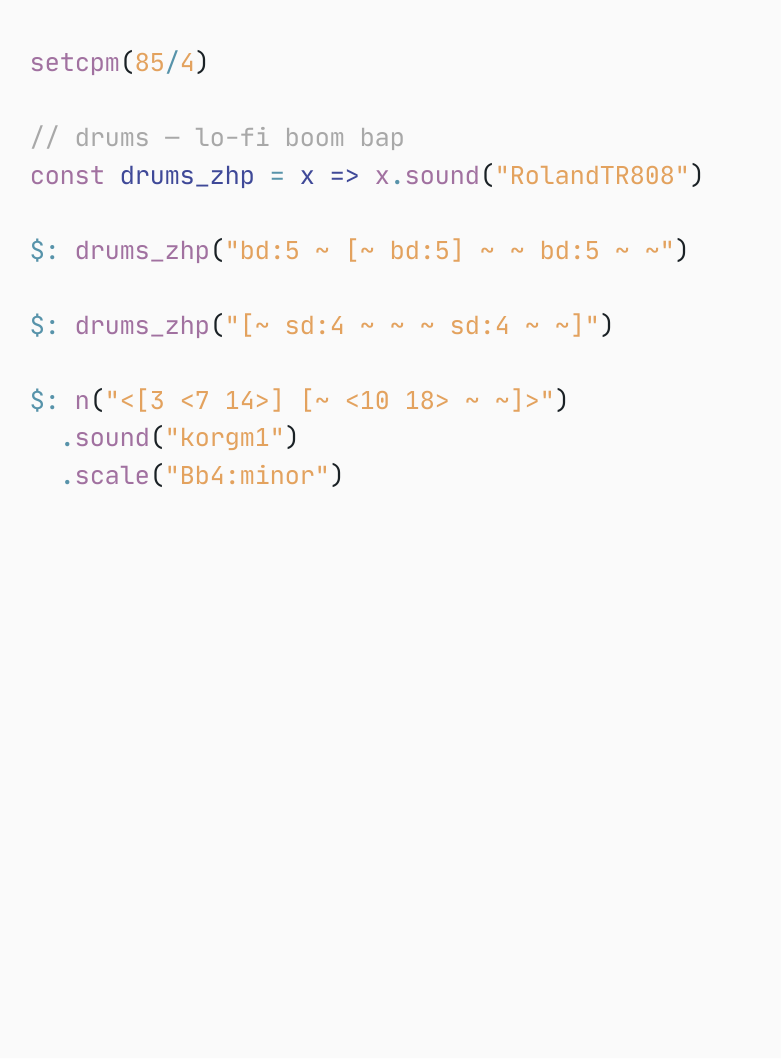}
        \caption{Qwen3-8B after SFT}
        \label{fig:apd:qual:sft}
    \end{subfigure}

    \caption{\textbf{Qualitative examples: trained model variants.} We show the input MIDI visualization and the corresponding \Strudel{} decompilations produced by \system{} (8B), the vanilla Qwen3-8B model, and Qwen3-8B after SFT. Sound-effect parameters such as {\prettyfont gain()} and {\prettyfont room()} are omitted so that the comparison focuses on how each method encodes the underlying MIDI structure.}
    \label{fig:apd:qual_examples_1}
\end{figure*}

%% file: figures/fig_apd_qual_examples_2.tex
\begin{figure*}[t!]
    \centering
    \begin{subfigure}[t]{0.49\linewidth}
        \centering
        \includegraphics[width=\linewidth]{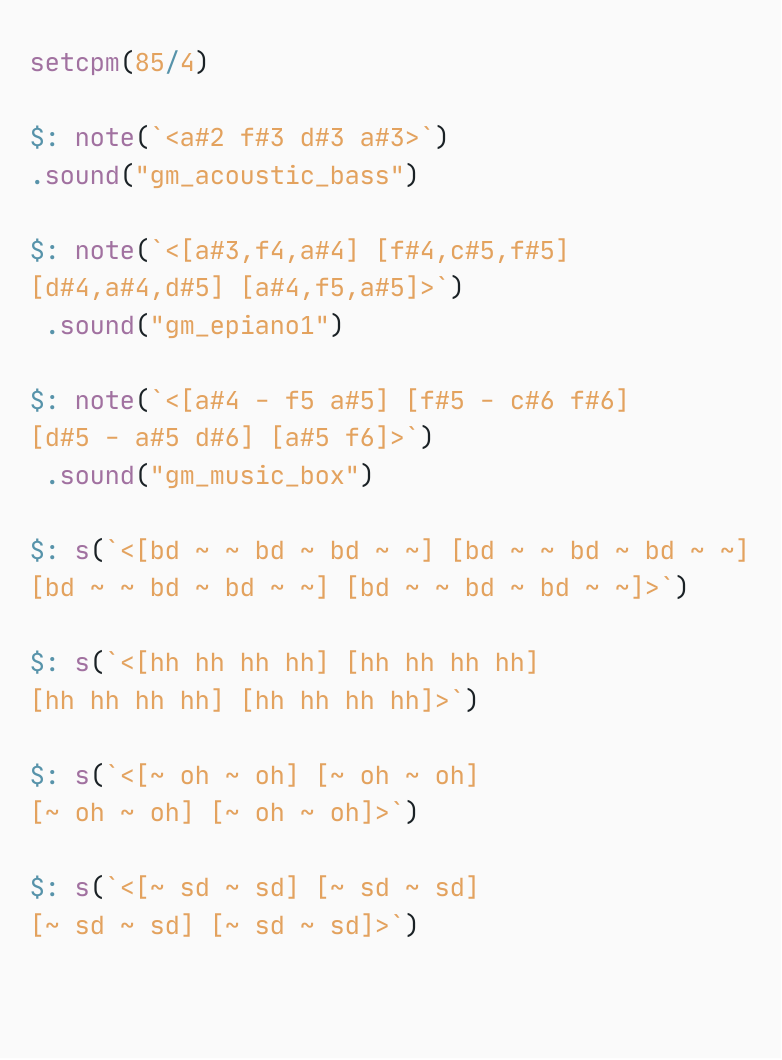}
        \caption{Heuristic converter}
        \label{fig:apd:qual:heuristic}
    \end{subfigure}
    \hfill
    \begin{subfigure}[t]{0.49\linewidth}
        \centering
        \includegraphics[width=\linewidth]{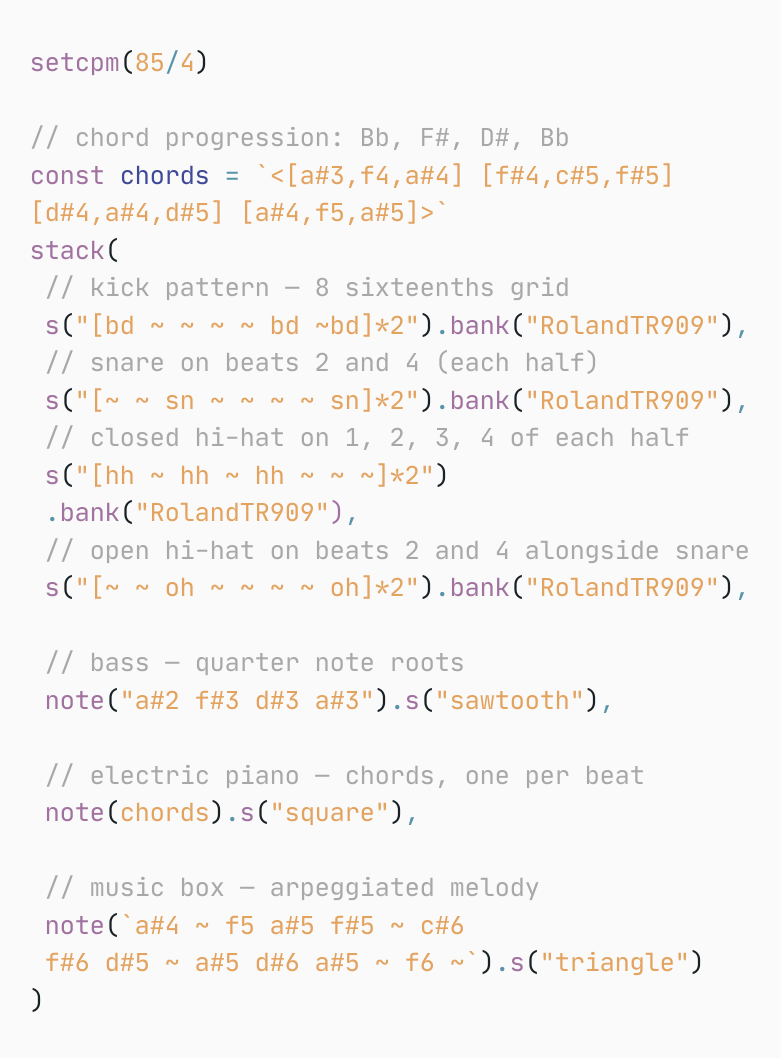}
        \caption{\claude{}}
        \label{fig:apd:qual:claude}
    \end{subfigure}

    \vspace{0.5em}

    \begin{subfigure}[t]{0.49\linewidth}
        \centering
        \includegraphics[width=\linewidth]{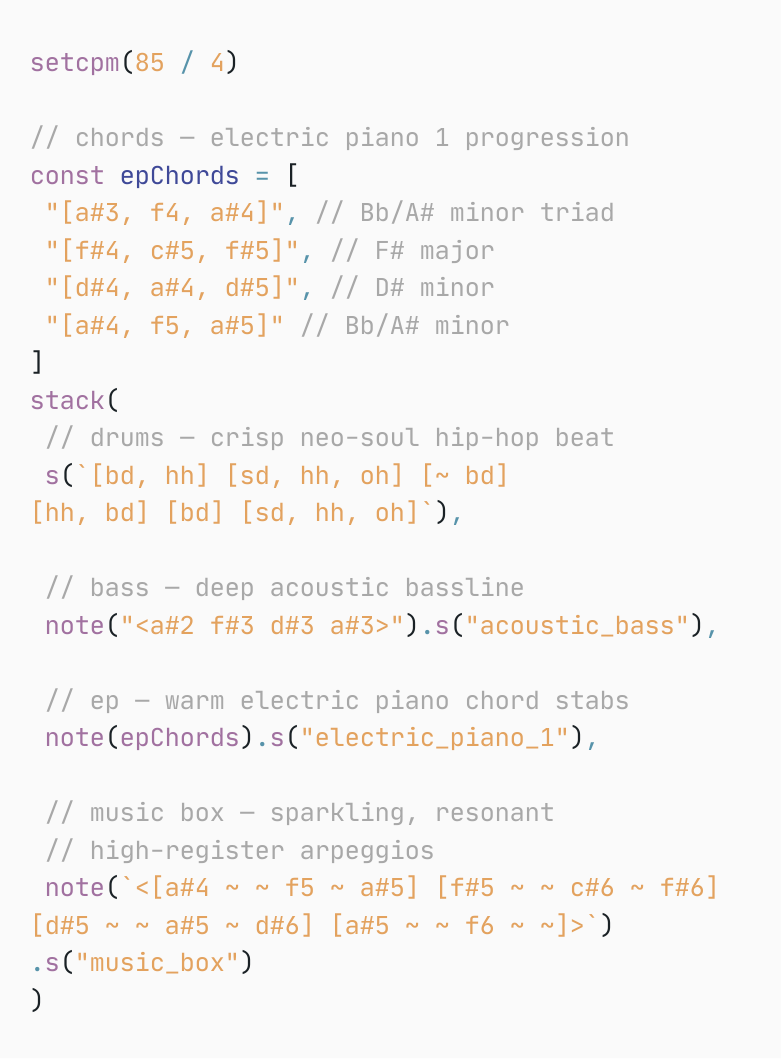}
        \caption{\gemini{}}
        \label{fig:apd:qual:gemini}
    \end{subfigure}
    \hfill
    \begin{subfigure}[t]{0.49\linewidth}
        \centering
        \includegraphics[width=\linewidth]{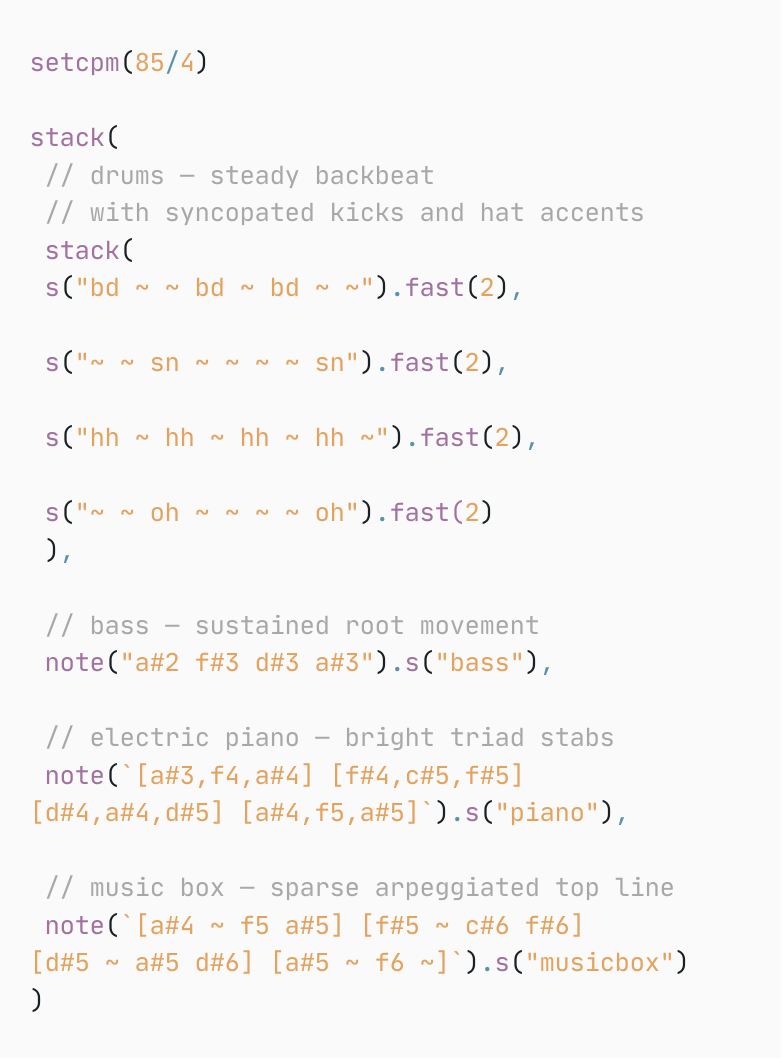}
        \caption{\gpt{}}
        \label{fig:apd:qual:gpt}
    \end{subfigure}

    \caption{\textbf{Qualitative examples: baselines.} For the same MIDI input as Figure~\ref{fig:apd:qual_examples_1}, we show decompilations produced by the heuristic converter and the frontier models (\claude{}, \gemini{}, and \gpt{}). Sound effect parameters are omitted for consistency.}
    \label{fig:apd:qual_examples_2}
\end{figure*}